\documentclass[10pt,twocolumn,letterpaper]{article}

\usepackage[pagenumbers]{wacv} 

\usepackage{booktabs}   
\usepackage{multirow}  
\usepackage{array}
\usepackage{color}
\usepackage{amssymb}
\usepackage{bbm}
\usepackage[table]{xcolor}
\usepackage{microtype}

\definecolor{wacvblue}{rgb}{0.21,0.49,0.74}
\usepackage[pagebackref,breaklinks,colorlinks,allcolors=wacvblue]{hyperref}

\def\wacvPaperID{2648} 
\def\confName{WACV}
\def\confYear{2026}
\newcommand{\lingqiao}[0]{\textcolor{black}}
\newcommand{\Ankit}[0]{\textcolor{black}}

\title{Revisiting Vision–Language Foundations for No-Reference Image Quality Assessment}

\author{
Ankit Yadav \quad Ta Duc Huy \quad Lingqiao Liu\\
\small Australian Institute for Machine Learning, The University of Adelaide, Australia\\
\tt\small \{ankit.yadav, huy.ta, lingqiao.liu\}@adelaide.edu.au
}

\begin{document}
\maketitle
\begin{abstract}
Large-scale vision–language pre-training has recently shown promise for no-reference image-quality assessment (NR-IQA), yet the relative merits of modern Vision Transformer foundations remain poorly understood. In this work, we present the first systematic evaluation of six prominent pretrained backbones, CLIP, SigLIP2, DINOv2, DINOv3, Perception, and ResNet, for the task of No-Reference Image Quality Assessment (NR-IQA), each fine-tuned using an identical lightweight MLP head. Our study uncovers two previously overlooked factors: (1) SigLIP2 consistently achieves strong performance; and (2) the choice of activation function plays a surprisingly crucial role, \lingqiao{ particularly for enhancing the generalization ability of image quality assessment models.} Notably, we find that simple sigmoid activations outperform commonly used ReLU and GELU on several benchmarks. Motivated by this finding, we introduce a learnable activation selection mechanism that adaptively determines the nonlinearity for each channel, eliminating the need for manual activation design, and
achieving new state-of-the-art SRCC on CLIVE, KADID10K, and AGIQA3K. Extensive ablations confirm the benefits across architectures and regimes, establishing strong, resource-efficient NR-IQA baselines.

\end{abstract}


\section{Introduction}

No-reference image-quality assessment (NR-IQA) estimates an image’s perceptual quality without access to a pristine reference. This task is pivotal for consumer photography, video streaming, and the burgeoning field of AI-generated imagery scenarios in which a ground-truth counterpart rarely exists. Yet NR-IQA remains challenging: mean-opinion scores (MOS) are noisy and costly to collect, distortion types are open-ended, and perceptual cues span low-level texture, high-level semantics, and device-specific artifacts \cite{bosse2017deep,zhang2018blind}.

Early deep‐learning approaches replaced hand-crafted features with convolutional regressors, WaDIQaM \cite{bosse2017deep} and DBCNN \cite{zhang2018blind} led the way, but their performance degraded when faced with unseen distortions or cross-dataset shifts. More recently, Vision Transformers (ViTs)\cite{dosovitskiy2020image} and large vision–language (VL) encoders such as CLIP\cite{radford2021learning} have demonstrated richer, more transferable representations for high-level tasks \cite{radford2021learning}. However, nearly all NR-IQA studies to date fine-tune a single backbone, typically CLIP, leaving open the question of how the choice of foundation model influences no-reference image-quality assessment performance.

\begin{figure}[t]  
  \centering
  \includegraphics[width=1\linewidth]{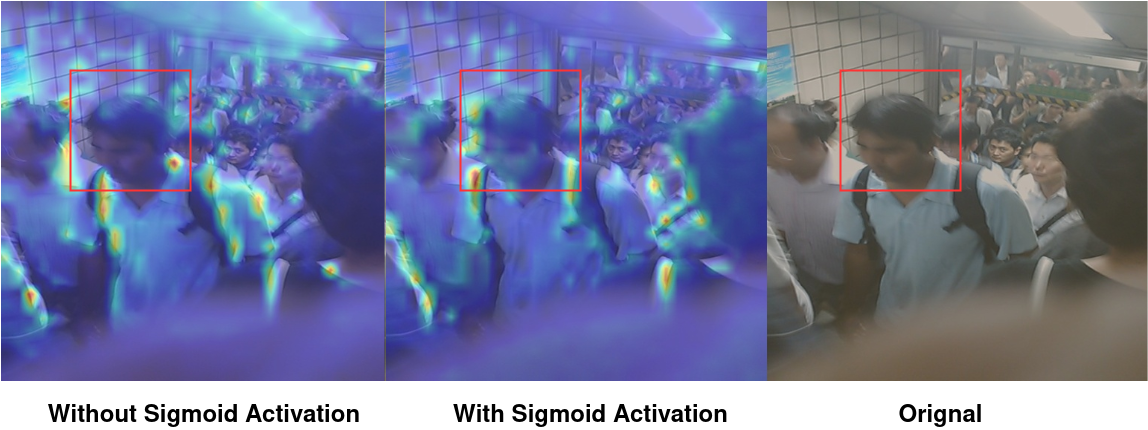}
  \caption{In this comparison, the sigmoid-activated MLP head provides more precise Grad-CAM localization of natural blur on and around faces, than the alternative variant, a cue closely tied to perceived image quality. Example from CLIVE\cite{ghadiyaram2015massive}.}
  \label{fig:sigmoid_vs_L_RELU_qualitative}
\end{figure}

In this work, we present the first systematic head-to-head evaluation of six leading pretrained encoders, CLIP\cite{radford2021learning}, SigLIP2\cite{tschannen2025siglip2multilingualvisionlanguage}, DINOv2\cite{oquab2023dinov2}, DINOv3\cite{simeoni2025dinov3}, Perception\cite{bolya2025perception}, and ResNet\cite{he2016deep}, each fine-tuned with an identical three-layer adapter via Low-Rank Adaptation (LoRA). Our study yields two key insights. First, SigLIP2-SO400M, consistently outperforms other backbones across both within- and cross-dataset settings. Second, we observe that the activation function used in the prediction head atop the pretrained encoder can markedly influence overall performance. Qualitative results in Figure~\ref{fig:sigmoid_vs_L_RELU_qualitative} further support this trend.
Hence, we introduce learnable activation functions that enable the model to adaptively select its nonlinear transformation, rather than relying on a fixed choice such as ReLU, GELU, or sigmoid.

When combined, these two innovations deliver new state-of-the-art Spearman rank correlation coefficients on CLIVE, KADID10K\cite{lin2020deepfl}, and AGIQA3K\cite{li2023agiqa} benchmarks. Extensive ablation experiments across architectures and training regimes confirm that dynamic activation selection 
yield complementary improvements, and establish resource-efficient baselines for future NR-IQA research.
Our contributions are as follows:
\begin{itemize}
    \item We conduct the first unified comparison of six foundation models for NR-IQA, uncovering the overlooked strength of SigLIP2-SO400M.
    \item We systematically quantify the effect of prediction-head activation functions on NR-IQA, demonstrating consistent gains from a sigmoid nonlinearity.
    \item We propose a learnable activation selection mechanism that adapts model nonlinearities to improve performance.
    \item Our method sets new state-of-the-art SRCC on CLIVE, KADID10K, and AGIQA3K, with an ablation study confirming the design choice.
\end{itemize}

\begin{table*}[!htbp]
\centering
\caption{Spearman (SRCC) and Pearson (PLCC) correlations for no‐reference IQA across seven datasets, averaged over three seeds (8,~19,~25). Standard deviations are reported in Table~\textbf{S2} in the supplementary. Each backbone is paired with a three‐layer MLP head and fine-tuned with \textbf{LoRA} adapters (rank~=~4). We report the baseline configuration (three-layer MLP with two interleaved LeakyReLU gates). The SigLIP2‐SO400M backbone outperforms the other encoders on most datasets. \textbf{Bold} indicates the best result for each configuration. \textbf{Note}: "Percept" refers to the perception encoder backbone. CDIFT refers to CleanDIFT with Stable Diffusion 2.1. "---" indicates experiments produced NAN for that dataset. 
\checkmark~indicates methods where the first activation layer is replaced with Sigmoid. (+Sig)}
\label{tab:different_backbone_performance_comparison}
\footnotesize
\setlength{\tabcolsep}{3pt}
\begin{tabular}{@{}lc*{14}{c}|*{2}{c}@{}}
\toprule
\multirow{2}{*}{Method} & \multirow{2}{*}{+Sig} & \multicolumn{2}{c}{FLIVE} & \multicolumn{2}{c}{SPAQ} & \multicolumn{2}{c}{CLIVE} & \multicolumn{2}{c}{AGIQA3K} & \multicolumn{2}{c}{KADID10K} & \multicolumn{2}{c}{KonIQ10K} & \multicolumn{2}{c|}{AGIQA1K} & \multicolumn{2}{c}{Average} \\
\cmidrule(lr){3-4} \cmidrule(lr){5-6} \cmidrule(lr){7-8} \cmidrule(lr){9-10} \cmidrule(lr){11-12} \cmidrule(lr){13-14} \cmidrule(lr){15-16} \cmidrule(lr){17-18}
& & SRCC & PLCC & SRCC & PLCC & SRCC & PLCC & SRCC & PLCC & SRCC & PLCC & SRCC & PLCC & SRCC & PLCC & SRCC & PLCC \\
\midrule
CLIP & & \textbf{.566} & \textbf{.676} & .916 & .919 & .841 & .872 & .841 & .896 & \textbf{.963} & \textbf{.966} & .884 & .906 & .820 & .870 & .833 & .872 \\
DINO2 & & .465 & .604 & .905 & .908 & .688 & .744 & .807 & .870 & .931 & .930 & .848 & .875 & .810 & .845 & .779 & .825 \\
DINO3 & & .546 & .663 & .914 & .918 & .750 & .801 & .843 & .900 & .939 & .938 & .904 & .920 & .812 & .860 & .815 & .857 \\
SigLIP2 & & .533 & .641 & \textbf{.927} & \textbf{.931} & \textbf{.875} & \textbf{.905} & \textbf{.865} & \textbf{.917} & .961 & .964 & \textbf{.932} & \textbf{.943} & \textbf{.857} & \textbf{.889} & \textbf{.850} & \textbf{.884} \\
Percept & & .472 & .556 & .912 & .913 & .812 & .845 & .853 & .903 & .957 & .957 & .881 & .897 & .847 & .882 & .819 & .850 \\
ResNet152 & & .388 & .401 & .815 & .815 & .533 & .528 & .623 & .690 & .522 & .552 & .287 & .220 & .656 & .688 & .546 & .556 \\
CDIFT & & .543 & .650 & .915 & .920 & .829 & .866 & .826 & .895 & .960 & .963 & .812 & .843 & .847 & .880 & .819 & .860 \\
\midrule
CLIP & \checkmark & .570 & .676 & .919 & .924 & .874 & .900 & .854 & .907 & .948 & .951 & .911 & .933 & .846 & .885 & .846 & .882 \\
DINO2 & \checkmark & .546 & .649 & \textbf{.922} & \textbf{.927} & .829 & .865 & .834 & .893 & .924 & .927 & .924 & .941 & .847 & .878 & .832 & .869 \\
DINO3 & \checkmark & \textbf{.581} & \textbf{.686} & .919 & .923 & .822 & .865 & .860 & .912 & .941 & .942 & .900 & .923 & .831 & .873 & .836 & .875 \\
SigLIP2 & \checkmark & .521 & .608 & .921 & .926 & \textbf{.909} & \textbf{.930} & \textbf{.878} & \textbf{.923} & .939 & .943 & \textbf{.938} & \textbf{.947} & \textbf{.872} & \textbf{.897} & \textbf{.854} & \textbf{.882} \\
Percept & \checkmark & .540 & .617 & .915 & .917 & .825 & .856 & .867 & .915 & .944 & .943 & .903 & .914 & .864 & .890 & .837 & .865 \\
ResNet152 & \checkmark & .465 & .517 & .854 & .848 & .661 & .688 & .678 & .796 & .689 & .677 & .767 & .773 & .731 & .781 & .692 & .726 \\
CDIFT & \checkmark & .541 & .621 & --- & --- & .835 & .869 & .829 & .893 & \textbf{.956} & \textbf{.960} & .855 & .880 & .856 & .885 & .812 & .851 \\
\bottomrule
\end{tabular}
\end{table*}

\section{Related Works}

Early no-reference IQA (NR‑IQA) methods relied on handcrafted statistical regularities of natural scenes either in the spatial domain, as in BRISQUE \cite{mittal2012no}
or frequency domains via log‑Gabor filter responses, as in ILNIQE \cite{zhang2015feature}.
Learning‑based approaches soon replaced fixed features with deep convolutional backbones, e.g.\ WaDIQaM’s patch MSE optimisation \cite{bosse2017deep}, QPT introduces a quality-aware contrastive objective and multi-degradation views to cluster patches by perceived quality rather than content \cite{zhao2023quality}, DBCNN’s dual‑stream design \cite{zhang2018blind}, and HyperIQA’s hyper‑network routing of weights \cite{su2020blindly}.
Despite impressive in‑dataset performance, these CNN models generalize poorly across distortion types or novel capture devices.

\noindent \textbf{Foundation Transformer Models in NR-IQA} Transformer backbones have driven NR‑IQA progress since~2021, when TIQA showed that a vanilla ViT can rival deeper CNNs on authentic distortions~\cite{you2021transformer}
Local Distortion Aware (LoDA) injects local‑distortion adapters into a frozen ViT to boost cross‑dataset robustness \cite{xu2024boosting}.
Concurrently, vision–language (VL) encoders pretrained with contrastive image-text objectives, most notably CLIP \cite{radford2021learning}, were adopted for IQA either by prompt engineering or lightweight heads \cite{tang2024clip}.
However, since each study employs different experimental settings and protocols, comparing the relative performance of different backbones becomes challenging.
Hence, in our work, we benchmark six heterogeneous VL foundation models, including \textbf{SigLIP2} sigmoid‑scaled contrastive pre‑training \cite{tschannen2025siglip2multilingualvisionlanguage},
\textbf{DINOv2} self‑distilled ViT \cite{oquab2023dinov2},
and \textbf{Perception} that unifies image, video, and 3‑D inputs \cite{bolya2025perception},
revealing a large, previously undocumented spread in baseline SRCC scores discussed in Table~\ref{tab:different_backbone_performance_comparison}.

Unconditional latent‑diffusion models provide rich multi‑scale features that correlate with human perception.
LGDM extracts denoising UNet activations and aligns them via perceptual‑consistency guidance, achieving state‑of‑the‑art scores on CLIVE and KonIQ \cite{saini2025lgdm}.
DP‑IQA pushes further by learning a small MLP atop time‑aggregated hyper‑features from Stable~Diffusion~2.1 \cite{fu2024dp},
while GenZIQA employs prompt‑conditioned diffusion priors to handle AI‑generated content \cite{de2024genziqa}.
Although powerful, diffusion pipelines incur heavy inference cost; Our approach sidesteps that overhead by pairing a \textbf{SigLIP2‑SO400M} encoder with an 800~K‑parameter head, yet still surpasses diffusion‑based methods on CLIVE, KADID10K, and AGIQA‑3K (Table~\ref{tab:main_results}). 

\begin{table*}[t]
\centering
\caption{Spearman (SRCC, SR) and Pearson (PLCC, PL) correlations for no-reference IQA across seven datasets using channel-wise gated MLP heads Section~\ref{sec:gated mlp analysis}. Each backbone is paired with a three-layer MLP that employs an adaptive activation Figure~\ref{fig:MLP-3-layer} and fine-tuned with \textbf{LoRA} adapters (rank~=~4). \textbf{Bold} indicates the best result per dataset. CDIFT refers to CleanDIFT with Stable Diffusion 2.1. Results are averaged over three seeds (8,~19,~25). Standard deviations are reported in Table S2 in the supplementary.}
\label{tab:gated_mlp_results}
\small 
\begin{tabular}{@{}l*{14}{c}|*{2}{c}@{}}
\toprule
\multirow{2}{*}{Method} & \multicolumn{2}{c}{SPAQ} & \multicolumn{2}{c}{FLIVE} & \multicolumn{2}{c}{CLIVE} & \multicolumn{2}{c}{AGIQA3K} & \multicolumn{2}{c}{KADID10K} & \multicolumn{2}{c}{KonIQ10K} & \multicolumn{2}{c|}{AGIQA1K} & \multicolumn{2}{c}{Average} \\
\cmidrule(lr){2-3} \cmidrule(lr){4-5} \cmidrule(lr){6-7} \cmidrule(lr){8-9} \cmidrule(lr){10-11} \cmidrule(lr){12-13} \cmidrule(lr){14-15} \cmidrule(lr){16-17}
& SR & PL & SR & PL & SR & PL & SR & PL & SR & PL & SR & PL & SR & PL & SR & PL \\
\midrule
CLIP       & .918 & .923 & \textbf{.575} & \textbf{.685} & .850 & .874 & .846 & .901 & .965 & .968 & .914 & .935 & .839 & .874 & .844 & .880 \\
DINO2       & .913 & .917 & .558 & .665 & .753 & .802 & .828 & .886 & .946 & .949 & .923 & .940 & .829 & .864 & .821 & .860 \\
DINO3      & .917 & .922 & .573 & .671 & .764 & .814 & .849 & .906 & .934 & .938 & .910 & .928 & .835 & .878 & .826 & .865 \\
SigLIP2    & \textbf{.928} & \textbf{.932} & .540 & .636 & \textbf{.887} & \textbf{.912} & \textbf{.869} & \textbf{.919} & \textbf{.970} & \textbf{.973} & \textbf{.953} & \textbf{.962} & \textbf{.873} & \textbf{.892} & \textbf{.860} & \textbf{.889} \\
Percept    & .911 & .914 & .528 & .603 & .826 & .855 & .861 & .908 & .954 & .954 & .891 & .906 & .854 & .881 & .832 & .860 \\
ResNet152  & .831 & .835 & .471 & .533 & .624 & .645 & .668 & .757 & .609 & .640 & .729 & .749 & .724 & .776 & .665 & .705 \\
CDIFT      & .916 & .921 & .543 & .643 & .830 & .866 & .831 & .897 & .958 & .961 & .898 & .920 & .841 & .877 & .831 & .869 \\
\bottomrule
\end{tabular}
\end{table*}

\begin{table}[t]
\centering
\caption{Cross-dataset NR-IQA. Each entry is the SRCC from training on one dataset and testing on another, averaged over three runs (seeds: 8, 19, 25) (Table~S4) \textbf{Bold} indicates the best SRCC and \underline{underline} indicates the second-best. We denote the student and teacher networks in DP-IQA by s and t, respectively, and our experiments as B: Baseline, B\_Sig: Baseline\_Sigmoid, B\_Gate: Baseline\_Gated. }
\label{tab:cross_dataset_results}
\resizebox{\columnwidth}{!}{%
\begin{tabular}{l|c|c|c|c|c}
\toprule
\textbf{Train set} & \textbf{CLIVE} & \textbf{KonIQ10K} & \textbf{FLIVE} & \textbf{FLIVE} & \textbf{Average} \\
\textbf{Test set} & \textbf{KonIQ10K} & \textbf{CLIVE} & \textbf{CLIVE} & \textbf{KonIQ10K} &  \\
\midrule
\textbf{Method} & \textbf{SRCC} & \textbf{SRCC} & \textbf{SRCC} & \textbf{SRCC} & \textbf{SRCC} \\
\midrule
DBCNN~\cite{zhang2018blind} & .754 & .755 & .724 & .716 & .737 \\
P2P-BM~\cite{ying2020patches} & .740 & .770 & .738 & .755 & .751 \\
HyperIQA~\cite{su2020blindly} & .772 & .785 & .735 & .758 & .762 \\
TReS~\cite{golestaneh2022no} & .733 & .786 & .740 & .713 & .743 \\
LoDa~\cite{xu2024boosting} & .745 & .811 & .805 & .763 & .781 \\
LGDM~\cite{saini2025lgdm} & .794 & \underline{.853} & \textbf{.849} & \underline{.802} & \textbf{.825} \\
DP-IQA~\cite{fu2024dp} & .781 (s) & .833 (T) & .770 (T) & .771 (T) & .789 \\
\midrule
B & .726 & .766 & \underline{.816} & .763 & .768 \\
B\_Sig (Ours) & \textbf{.808} & .820 & .796 & \textbf{.807} & .808 \\
B\_Gated (Ours) & \underline{.804} & \textbf{.899} & .786 & .796 & \underline{.821} \\
\bottomrule
\end{tabular}
}
\end{table}


\begin{figure*}[t]  
  \centering
  \includegraphics[width=0.99\linewidth]{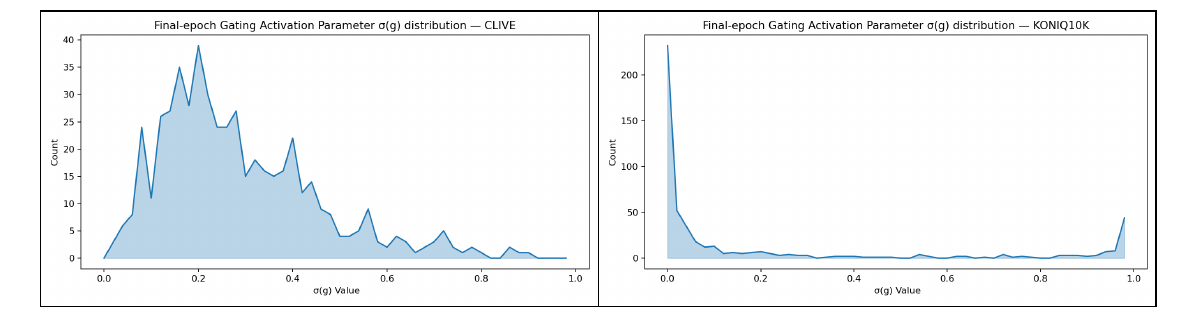}
  \caption{Channel-wise distributions of the gate weight $w = \sigma(g)$ learned by the gated activation head for the final epoch, comparing CLIVE (low-data regime) and KonIQ10K (large-data regime). Larger $w$ indicates greater reliance on the sigmoid branch, while smaller $w$ favors the LeakyReLU branch. See supplementary Figure~S7 For epoch-wise distribution. }
  \label{fig: Channel Wise distribution}
\end{figure*}

\begin{table*}[t]
\centering
\caption{Comparison of MLP performance under different feature-retention percentiles $k$ \emph{(masking variants)}. Reported are the mean best Spearman (SRCC) and Pearson (PLCC) for Group~1 (synthetic: AGIQA1K, KADID10K) and Group~2 (natural: CLIVE, KonIQ10K). $\Delta$ denotes the change relative to $k=100$ within each group (more negative indicates a larger drop). Here, $k$ is the retained percentile of features by magnitude used in training/evaluation. We conduct all experiments with a SigLIP2 backbone.}
\label{tab:sampling_ratio_results_grouped}
\small
\begin{tabular}{@{}lc*{4}{c}|*{2}{c}|*{4}{c}|*{2}{c}@{}}
\toprule
\multirow{2}{*}{Method} & \multirow{2}{*}{k} 
& \multicolumn{2}{c}{AGIQA1K} & \multicolumn{2}{c}{KADID10K} 
& \multicolumn{2}{|c|}{$\Delta$ Group 1} 
& \multicolumn{2}{c}{CLIVE} & \multicolumn{2}{c}{KonIQ10K} 
& \multicolumn{2}{|c}{$\Delta$ Group 2} \\
\cmidrule(lr){3-4} \cmidrule(lr){5-6} \cmidrule(lr){7-8} \cmidrule(lr){9-10} \cmidrule(lr){11-12} \cmidrule(lr){13-14}
& & SRCC & PLCC & SRCC & PLCC & SRCC & PLCC & SRCC & PLCC & SRCC & PLCC & SRCC & PLCC \\
\midrule

\multirow{6}{*}{\parbox{2cm}{MLP-LReLU\\(Masking)}} 
& 100 & .852 & .886 & .883 & .885 & 0 & 0 & .866 & .902 & .910 & .918 & 0 & 0 \\
& 90  & .837 & .870 & .864 & .865 & -0.017 & -0.018 & .844 & .883 & .896 & .896 & -0.018 & -0.021 \\
& 70  & .793 & .852 & .837 & .830 & -0.052 & -0.044 & .834 & .873 & .855 & .868 & -0.043 & -0.050 \\
& 50  & .826 & .864 & .836 & .836 & -0.037 & -0.035 & .846 & .871 & .846 & .818 & -0.042 & -0.052 \\
& 30  & .762 & .818 & .610 & .619 & -0.182 & -0.167 & .780 & .786 & .502 & .545 & -0.162 & -0.190 \\
& 10  & .580 & .539 & .312 & .328 & -0.421 & -0.402 & .451 & .451 & .388 & .415 & -0.423 & -0.454 \\
\midrule

\multirow{6}{*}{\parbox{2cm}{MLP-Sigmoid\\(Masking)}} 
& 100 & .865 & .895 & .850 & .855 & 0 & 0 & .898 & .924 & .906 & .923 & 0 & 0 \\
& 90  & .852 & .888 & .835 & .839 & -0.012 & -0.012 & .888 & .916 & .895 & .908 & -0.006 & -0.011 \\
& 70  & .829 & .878 & .774 & .777 & -0.031 & -0.048 & .886 & .905 & .902 & .913 & -0.008 & -0.014 \\
& 50  & .817 & .863 & .701 & .704 & -0.066 & -0.092 & .879 & .898 & .777 & .793 & -0.024 & -0.036 \\
& 30  & .734 & .797 & .527 & .519 & -0.227 & -0.217 & .738 & .767 & .605 & .625 & -0.090 & -0.118 \\
& 10  & .326 & .314 & .103 & .105 & -0.595 & -0.586 & .235 & .247 & .148 & .149 & -0.470 & -0.475 \\
\bottomrule
\end{tabular}
\end{table*}



Several works improve data efficiency by dispensing with MOS labels.
ARNIQA learns a “distortion manifold’’ through SimCLR on synthetically degraded pairs and trains only a linear regressor for scoring \cite{agnolucci2024arniqa}.
Re‑IQA mixes quality‑aware and content‑aware encoders under a mutual‑learning scheme to reach competitive zero‑shot performance \cite{saha2023re}
MetaIQA instead meta‑trains across distortion families so that few‑shot fine‑tuning suffices on new domains \cite{zhu2020metaiqa}.
Our approach remains supervised but shows that judicious architectural tweaks like sigmoid activation and parameterized nonlinearities yield larger gains than elaborate training curricula.

\noindent \textbf{Activation Functions and regularization in IQA Heads}
Most prior IQA heads adopt \textbf{ReLU} or \textbf{GELU} without in-depth justification.
TReS~\cite{golestaneh2022no} reports marginal benefits from Swish but focuses on self‑consistency losses \cite{golestaneh2022no}; no systematic study of activations has been conducted.
Likewise, regularisers have targeted ranking consistency or distortion‑aware adapters \cite{xu2024boosting}, yet leave the feature geometry largely unconstrained.
To our knowledge, this is the first work to
(i) show that replacing a single ReLU with a Sigmoid raises SRCC by up to 3 percentage points across multiple benchmarks in low-data settings;
(ii) propose a three-layer MLP head with channel-wise gating and a learnable non-linearity that improves performance in both low- and high-data regimes. 

In summary, existing efforts either specialise in one backbone, demand heavy diffusion inference, or introduce elaborate training schedules.
Our study fills this gap by \emph{revisiting} VL foundation models under a unified fine‑tuning recipe, uncovering the underrated role of activation choice, and proposing a learnable activation selection mechanism that achieves state‑of‑the‑art NR‑IQA results. 

\section{Impact of the foundation models on NR-IQA}
\label{sec:foundation-models}

Recent studies have begun to tap large vision--language (VL) encoders and diffusion backbones for NR-IQA, e.g., \ CLIP‑IQA, DP‑IQA, and LGDM. Where the CLIP-IQA is based on CLIP backbone, while DP-IQA and LGDM leverage a diffusion backbone like (Stable Diffusion)SD~2.1 or SD~1.5 \cite{Rombach_2022_CVPR}, yet almost all fixate on a \emph{single} backbone, leaving open how much the \emph{choice} of encoder itself shapes NR‑IQA. 
Hence, we address this gap by systematically comparing pretrained backbones.

\subsection{Experiment Setup}
\label{sec:exp-setup}

We therefore benchmark six diverse open-source foundation backbones SigLIP2‑SO400m~\cite{tschannen2025siglip2multilingualvisionlanguage}, CLIP‑ViT‑L/14 \cite{radford2021learning}, DINOv2‑Large~\cite{oquab2023dinov2}, DINOv3-ViT-H/16~\cite{simeoni2025dinov3}, Perception‑ViT‑L14‑336~\cite{bolya2025perception}, and ResNet‑152~\cite{he2016deep} under an identical three-layer MLP adapter with a LeakyReLU activation after the first and second layers and LoRA adapters for the backbones. We will refer to this configuration as the \textbf{Baseline} in the rest of the paper. We also test a CleanDIFT‑tuned SD‑2.1 encoder  ~\cite {stracke2025cleandift} with a similar MLP and LoRA setup for completeness. The results are depicted in the Table~\ref{tab:different_backbone_performance_comparison}. 

We conduct our experiments on following public benchmarks where \textbf{CLIVE} offers 1\,162 in‑the‑wild photos rated on a 0--100 MOS scale~\cite{ghadiyaram2015massive}, \textbf{KonIQ10K} extends to 10\,073 crowd‑scored images~\cite{hosu2020koniq}, \textbf{KADID10K} provides 10\,125 synthetically distorted images across 25 distortion types~\cite{lin2020deepfl}, \textbf{FLIVE} contains 39\,810 social‑media photos with 4\,M ratings~\cite{ying2020patches}, \textbf{SPAQ} focuses on 11\,125 smartphone pictures~\cite{fang2020perceptual}, while \textbf{AGIQA1k}~\cite{zhang2023perceptual} and \textbf{AGIQA3k} target AI‑generated images~\cite{li2023agiqa}

We also ablate the importance of backbone fine-tuning by comparing three settings: (i) a frozen backbone, (ii) backbone fine-tuning via LoRA adapters, and (iii) full end-to-end fine-tuning. In all the experiments, we have the same backbone as SigLIP2‑SO400m and the baseline MLP setup discussed below. We observe that adding LORA adapters during training improves the performance in general
, as depicted in Table~\textbf{S1} (supplementary).

Each backbone is fine‑tuned with a lightweight \emph{LoRA} adapter that inserts rank‑4 update matrices into the query and key projections; we set the LoRA scaling factor to~8 and apply a 0.05 dropout during training~\cite{hu2022lora}.  Unless stated otherwise, all experiments run for 30\, epochs with Adam optimizer and a base learning rate of $1\times10^{-4}$.  
For the three small‑scale datasets (CLIVE, AGIQA1K, AGIQA3K), the learning rate remains constant, whereas for the four large‑scale datasets (KonIQ10K, KADID10K, FLIVE, SPAQ) we employ a \texttt{MultiStepLR} scheduler that multiplies the rate by~0.2 at epochs~15 and~25. 
Images are resized to $512$ size and pre‑processed with the native recipe of each encoder.  

\subsection{Optimisation Objective}
\label{sec:loss}

The network is trained with mean‑squared error augmented by the pair‑wise margin ranking term as discussed in Eq.~\ref{eq:margin}, and we report SRCC and PLCC averaged over three random train/test splits to mitigate variance.  
To encourage ordinal consistency, we add the pair‑wise margin loss (\textbf{Margin ranking loss}) with MSE as discussed in work \cite{fu2024dp}
\begin{equation}
  \mathcal{L}_{\text{margin}}
  =\frac{2}{n(n-1)}\!\sum_{i<j}\!
     \max\bigl\{0,\; -\operatorname{sgn}(s_i-s_j)(\hat{s}_i-\hat{s}_j)+m\bigr\},
  \label{eq:margin}
\end{equation}
where $\hat{s}$ are predictions, $m=\lambda_{\!m}\,\sigma_y$ is a dynamic
margin proportional to the ground‑truth standard deviation, $\sigma_y$ is the standard deviation of ground truth
($\lambda_{\!m}=0.25$), and $n$ is the batch size.

\begin{equation}
  \mathcal{L} \;=\; \mathcal{L}_{\text{MSE}} + \mathcal{L}_{\text{margin}} .
  \label{eq:total}
\end{equation}
This composite objective encourages both point‑wise accuracy and ordinal consistency without introducing additional hyper‑parameters beyond the fixed $\lambda_{\!m}=0.25$ used in Eq.~\ref{eq:margin}.

\subsection{Observations}
Table~\ref{tab:different_backbone_performance_comparison} shows that \textbf{encoder choice dominates performance}: SigLIP2 offers a strong baseline that already achieves \textbf{0.875} mean SRCC across the CLIVE dataset, surpassing Re-IQA \cite{saha2023re} \textbf{0.840} and MUSIQ \cite{ke2021musiq} \textbf{0.702} on CLIVE (Table~\ref{tab:main_results}a) and rivaling diffusion‑heavy LGDM on SPAQ (Table~\ref{tab:main_results}b). CLIP and Perception trail by 6 to 7 ~points on CLIVE; DINOv2 and ResNet‑152 lag further (Table~\ref{tab:different_backbone_performance_comparison}), confirming the benefits of contrastive VL pre‑training for perceptual regression over purely self‑supervised features. Freezing the backbone reduces performance by roughly 3 – 20 SRCC points, depending on the dataset, while full fine-tuning offers comparable performance to LoRA at a higher computational cost, underscoring the need for LoRA-based fine-tuning (see Table~\textbf{S1} in the supplementary.)

\lingqiao{Our results suggest that vision–language encoders such as SigLIP2 outperform purely visual self-supervised encoders (DINOv2, DINOv3) and CNN backbones (ResNet-152) largely because their contrastive image–text pretraining exposes them to a much broader and semantically richer visual distribution. Unlike traditional NR-IQA models that rely on low-level texture cues or distortion-specific patterns, VL foundations learn to align visual representations with high-level semantic concepts described in natural language. This alignment encourages them to encode both local perceptual details and global semantic structure, which appear to be crucial for judging perceptual quality in unconstrained real-world images where the quality degradation often interacts with semantics (e.g., faces, objects, and fine textures).}
\lingqiao{
 This finding has two major implications. First, it reframes NR-IQA not merely as a low-level perceptual regression problem but as a semantic–perceptual reasoning task: models must detect whether semantically important content is preserved under degradations. Second, it implies that future NR-IQA approaches could benefit from leveraging multimodal pretraining signals, such as caption-based supervision or cross-modal consistency losses, to further bridge the semantic gap between human quality perception and pixel-level distortions. As VL foundations become more powerful and efficient, they offer a promising path toward data-efficient, domain-robust IQA systems that generalize across capture devices, content domains, and generative models.}


\begin{figure}[t]  
  \centering
  \includegraphics[width=0.99\linewidth]{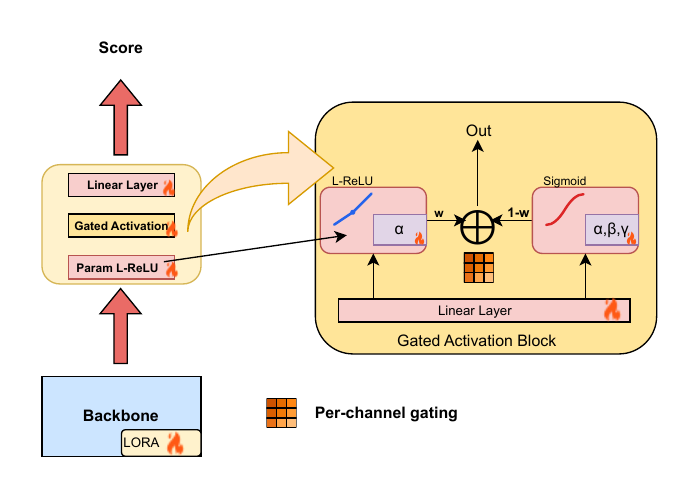}
  \caption{This figure depicts our Adaptive Gated MLP where both the activation layers of the network consist of parameterized Leaky-ReLU and Sigmoid whose outputs are mixed per channel through a learnable gate $(w_{c} = \sigma(g_{c}))$. All activation parameters are learned jointly with the linear weights.}
  \label{fig:MLP-3-layer}
\end{figure}

\begin{table*}[t]
\centering
\scriptsize
\setlength{\tabcolsep}{2pt}
\caption{Performance comparison with state-of-the-art methods on seven benchmark datasets. Best and second-best results are highlighted in \textbf{bold} and \underline{underlined}, respectively. B: Baseline, B\_Sig: Baseline\_Sigmoid, B\_Gate: Baseline\_Gated ,
Values represent Spearman (SRCC) and Pearson (PLCC) correlations averaged over three runs (seeds: 8, 19, 25). Standard deviations are reported in Table~\textbf{S2} in the supplementary.}
\label{tab:main_results}

\begin{subtable}[t]{\textwidth}
\centering
\subcaption{Non-diffusion methods}
\begin{tabular}{lcccccccccccccccc}
\toprule
\multirow{2}{*}{Methods} & \multicolumn{2}{c}{CLIVE} & \multicolumn{2}{c}{KonIQ10K} & \multicolumn{2}{c}{FLIVE} & \multicolumn{2}{c}{SPAQ} & \multicolumn{2}{c}{AGIQA3K} & \multicolumn{2}{c}{AGIQA1K} & \multicolumn{2}{c}{KADID10K} & \multicolumn{2}{c}{Average} \\
\cmidrule(lr){2-3} \cmidrule(lr){4-5} \cmidrule(lr){6-7} \cmidrule(lr){8-9} \cmidrule(lr){10-11} \cmidrule(lr){12-13} \cmidrule(lr){14-15} \cmidrule(lr){16-17}
 & SRCC↑ & PLCC↑ & SRCC↑ & PLCC↑ & SRCC↑ & PLCC↑ & SRCC↑ & PLCC↑ & SRCC↑ & PLCC↑ & SRCC↑ & PLCC↑ & SRCC↑ & PLCC↑ & SRCC↑ & PLCC↑ \\
\midrule
ILNIQE~\cite{zhang2015feature} & 0.508 & 0.508 & 0.523 & 0.537 & - & - & 0.713 & 0.712 & - & - & - & - & 0.534 & 0.558 & 0.570 & 0.579 \\
BRISQUE~\cite{mittal2012no} & 0.629 & 0.629 & 0.681 & 0.685 & 0.303 & 0.341 & 0.809 & 0.817 & - & - & - & - & 0.528 & 0.567 & 0.590 & 0.608 \\
WaDIQaM~\cite{bosse2017deep} & 0.682 & 0.671 & 0.804 & 0.807 & 0.455 & 0.467 & - & - & - & - & - & - & 0.739 & 0.752 & 0.670 & 0.674 \\
DBCNN~\cite{zhang2018blind} & 0.851 & 0.869 & 0.875 & 0.884 & 0.545 & 0.551 & 0.911 & 0.915 & - & - & - & - & 0.851 & 0.856 & 0.807 & 0.815 \\
TIQA~\cite{you2021transformer} & 0.845 & 0.861 & 0.892 & 0.903 & 0.541 & 0.581 & - & - & - & - & - & - & 0.850 & 0.855 & 0.782 & 0.800 \\
MetaIQA~\cite{zhu2020metaiqa} & 0.835 & 0.802 & 0.887 & 0.856 & 0.540 & 0.507 & - & - & - & - & - & - & 0.762 & 0.775 & 0.756 & 0.735 \\
P2P-BM~\cite{ying2020patches} & 0.844 & 0.842 & 0.872 & 0.885 & 0.526 & 0.598 & - & - & - & - & - & - & 0.840 & 0.849 & 0.770 & 0.793 \\
HyperIQA~\cite{su2020blindly} & 0.859 & 0.882 & 0.906 & 0.917 & 0.544 & 0.602 & 0.911 & 0.915 & - & - & - & - & 0.852 & 0.845 & 0.814 & 0.832 \\
TReS~\cite{golestaneh2022no} & 0.846 & 0.877 & 0.915 & 0.928 & 0.554 & 0.625 & - & - & - & - & - & - & 0.859 & 0.859 & 0.794 & 0.822 \\
MUSIQ~\cite{ke2021musiq} & 0.702 & 0.746 & 0.916 & 0.928 & 0.566 & 0.661 & 0.918 & 0.921 & - & - & - & - & 0.875 & 0.872 & 0.795 & 0.826 \\
CONTRIQUE~\cite{madhusudana2022image} & - & - & - & - & - & - & - & - & 0.804 & 0.868 & 0.670 & 0.708 & - & - & 0.737 & 0.788 \\
RE-IQA~\cite{saha2023re} & 0.840 & 0.854 & 0.914 & 0.923 & \textbf{0.645} & \textbf{0.733} & 0.918 & 0.925 & 0.785 & 0.845 & 0.614 & 0.670 & 0.872 & 0.885 & 0.798 & 0.834 \\
GenZIQA~\cite{de2024genziqa} & - & - & - & - & - & - & - & - & 0.832 & 0.892 & 0.840 & 0.861 & - & - & 0.836 & 0.877 \\
LoDA~\cite{xu2024boosting} & 0.876 & 0.899 & 0.932 & 0.944 & 0.578 & 0.679 & 0.925 & 0.928 & - & - & - & - & 0.931 & 0.936 & 0.848 & 0.877 \\
QCN~\cite{shin2024blind} & 0.875 & 0.893  & 0.934 & 0.945 & \underline{0.644} & \underline{0.741} & 0.923 & 0.928 & - & - & - & - & -& - & - & - \\
\midrule
B & 0.875 & 0.905 & 0.932 & 0.943 & 0.533 & 0.641 & \underline{0.927} & \underline{0.931} & 0.865 & 0.917 & 0.857 & 0.889 & \underline{0.961} & \underline{0.964} & 0.850 & 0.884 \\
B\_Sig\ (Ours) & \textbf{0.909} & \textbf{0.930} & \underline{0.938} & \underline{0.947} & 0.521 & 0.608 & 0.921 & 0.926 & \textbf{0.878} & \textbf{0.923} & \underline{0.872} & \textbf{0.897} & 0.939 & 0.943 & 0.854 & 0.882 \\
B\_Gated\ (Ours)  & \underline{0.887} & \underline{0.912} & \textbf{0.953} & \textbf{0.962} & 0.556 & 0.647 & \textbf{0.928} & \textbf{0.932} & \underline{0.867} & \underline{0.919} & \textbf{0.873} & \underline{0.892} & \textbf{0.970} & \textbf{0.973} & \textbf{0.862} & \textbf{0.891} \\
\bottomrule
\end{tabular}
\end{subtable}

\vspace{0.5em}

\begin{subtable}[t]{\textwidth}
\centering
\subcaption{Diffusion methods}
\begin{tabular}{lcccccccccccccccc}
\toprule
\multirow{2}{*}{Methods} & \multicolumn{2}{c}{CLIVE} & \multicolumn{2}{c}{KonIQ10K} & \multicolumn{2}{c}{FLIVE} & \multicolumn{2}{c}{SPAQ} & \multicolumn{2}{c}{AGIQA3K} & \multicolumn{2}{c}{AGIQA1K} & \multicolumn{2}{c}{KADID10K} & \multicolumn{2}{c}{Average} \\
\cmidrule(lr){2-3} \cmidrule(lr){4-5} \cmidrule(lr){6-7} \cmidrule(lr){8-9} \cmidrule(lr){10-11} \cmidrule(lr){12-13} \cmidrule(lr){14-15} \cmidrule(lr){16-17}
 & SRCC↑ & PLCC↑ & SRCC↑ & PLCC↑ & SRCC↑ & PLCC↑ & SRCC↑ & PLCC↑ & SRCC↑ & PLCC↑ & SRCC↑ & PLCC↑ & SRCC↑ & PLCC↑ & SRCC↑ & PLCC↑ \\
\midrule
DP-IQA~\cite{fu2024dp} & 0.893 & 0.913 & 0.942 & 0.951 & \underline{0.579} & \underline{0.683} & 0.923 & 0.926 & - & - & - & - & - & - & 0.834 & 0.868 \\
LGDM~\cite{saini2025lgdm} & \underline{0.908} & \textbf{0.940} & \textbf{0.967} & \textbf{0.972} & \textbf{0.705} & \textbf{0.812} & \textbf{0.947} & \textbf{0.948} & 0.863 & \textbf{0.929} & \textbf{0.891} & \textbf{0.903} & 0.958 & 0.961 & \textbf{0.891} & \textbf{0.924} \\
\midrule
B & 0.875 & 0.905 & 0.932 & 0.943 & 0.533 & 0.641 & 0.927 & 0.931 & 0.865 & 0.917 & 0.857 & 0.889 & \underline{0.961} & \underline{0.964} & 0.850 & 0.884 \\
B\_Sig (Ours) & \textbf{0.909} & \underline{0.930} & 0.938 & 0.947 & 0.521 & 0.608 & 0.921 & 0.926 & \textbf{0.878} & \underline{0.923} & 0.872 & \underline{0.897} & 0.939 & 0.943 & 0.854 & 0.882 \\
B\_Gated (Ours)  & 0.887 & 0.912 & \underline{0.953} & \underline{0.962} & 0.556 & 0.647 & \underline{0.928} & \underline{0.932} & \underline{0.867} & 0.919 & \underline{0.873} & 0.892 & \textbf{0.970} & \textbf{0.973} & \underline{0.862} & \underline{0.891} \\
\bottomrule
\end{tabular}
\end{subtable}
\end{table*}

\section{Activation Function in MLP Matters}

\begin{table}[t]
\small
\setlength{\tabcolsep}{3.5pt}
\caption{Activation-function ablation in our three-layer MLP head (Act$_1$ for the first hidden layer, Act$_2$ for the second), trained with MSE loss. We report Spearman rank-correlation (SRCC) and Pearson linear-correlation (PLCC), respectively.  \textbf{Bold} highlights the best score in each column. All the experiments are conducted on SigLIP2 Backbone.}
\label{tab:activation_function_comparison}
\resizebox{\columnwidth}{!}{%
\begin{tabular}{lcccccc|cc}
\toprule
\multirow{2}{*}{\textbf{Act1 + Act2}} & \multicolumn{2}{c}{\textbf{AGIQA1K}} & \multicolumn{2}{c}{\textbf{CLIVE}} & \multicolumn{2}{c|}{\textbf{AGIQA3K}} & \multicolumn{2}{c}{\textbf{Average}} \\
\cmidrule(lr){2-3} \cmidrule(lr){4-5} \cmidrule(lr){6-7} \cmidrule(lr){8-9}
 & SRCC & PLCC & SRCC & PLCC & SRCC & PLCC & SRCC & PLCC \\
\midrule
GELU + GELU & .869 & .894 & .881 & .908 & .881 & .925 & .877 & .909 \\
GELU + LReLU & .868 & .897 & .872 & .907 & .880 & .925 & .873 & .910 \\
LReLU + LReLU & .866 & .891 & .870 & .908 & .875 & .922 & .870 & .907 \\
Sig + LReLU & \textbf{.870} & \textbf{.898} & .900 & .923 & \textbf{.882} & \textbf{.928} & \textbf{.884} & .916 \\
Sig + GELU & .866 & .895 & \textbf{.910} & \textbf{.931} & .877 & .924 & \textbf{.884} & \textbf{.917} \\
Tanh + Tanh & .866 & .895 & .874 & .902 & .879 & .922 & .873 & .906 \\
Tanh + LReLU & .869 & .898 & .869 & .903 & .879 & .922 & .872 & .908 \\
\bottomrule
\end{tabular}
}
\end{table}

\label{sec: Activation function Matters}

\lingqiao{The experiments in Section 3 showed that the choice of vision–language foundation backbone has a substantial impact on NR-IQA performance, with SigLIP2 emerging as the strongest overall. While this establishes the value of stronger encoders, the prediction head atop these backbones remains largely unexplored, and most prior NR-IQA studies adopt standard ReLU-family activations (e.g. LeakyReLU, GELU) without justification. Given that the head is directly responsible for mapping rich semantic features into perceptual quality scores, its nonlinearity could critically shape what information is preserved or suppressed. In this section, we therefore investigate the role of activation functions in the prediction head.}
\begin{figure*}[t]  
  \centering
  \includegraphics[width=.48\linewidth]{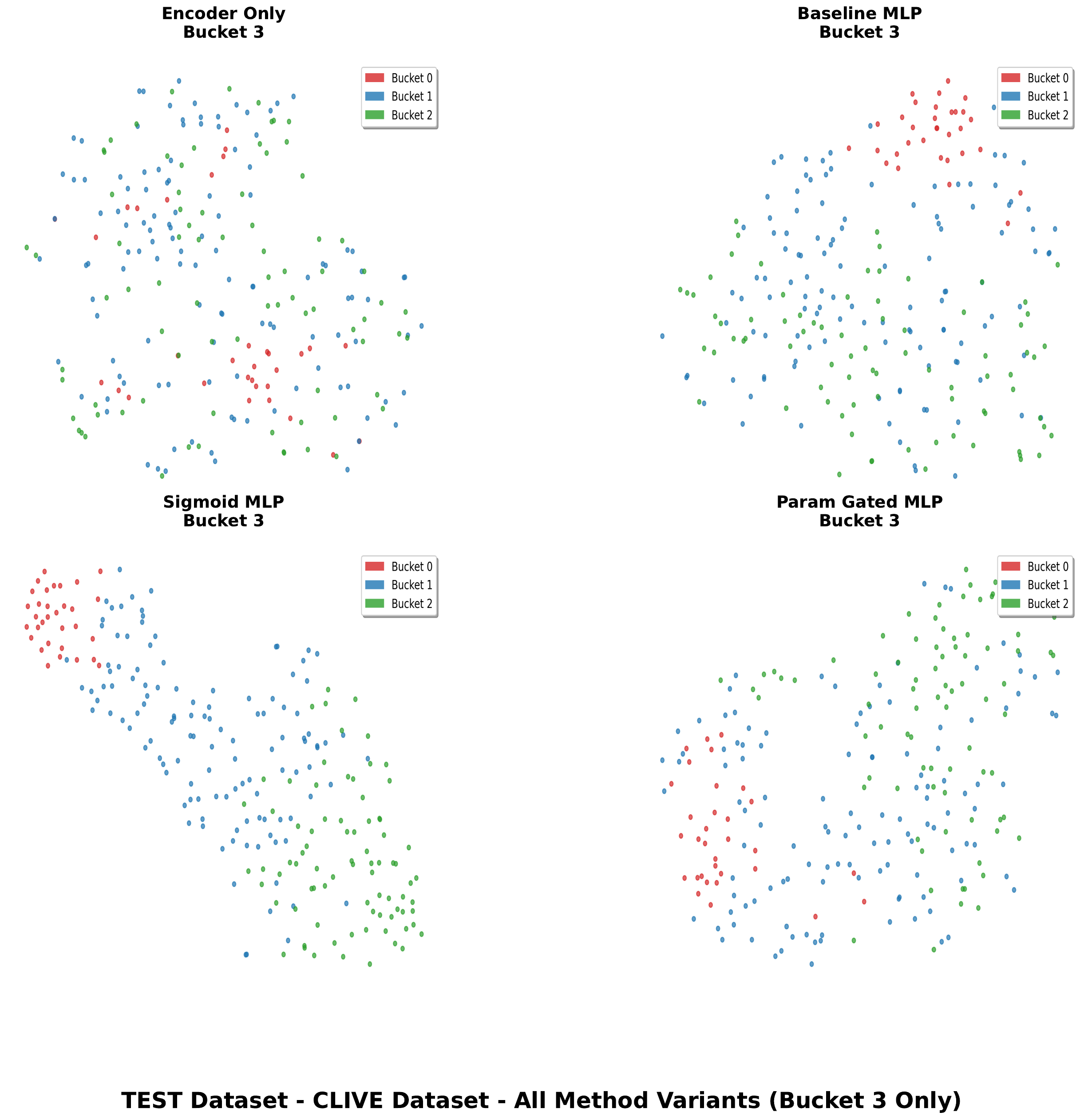}\hfill
  \includegraphics[width=.48\linewidth]{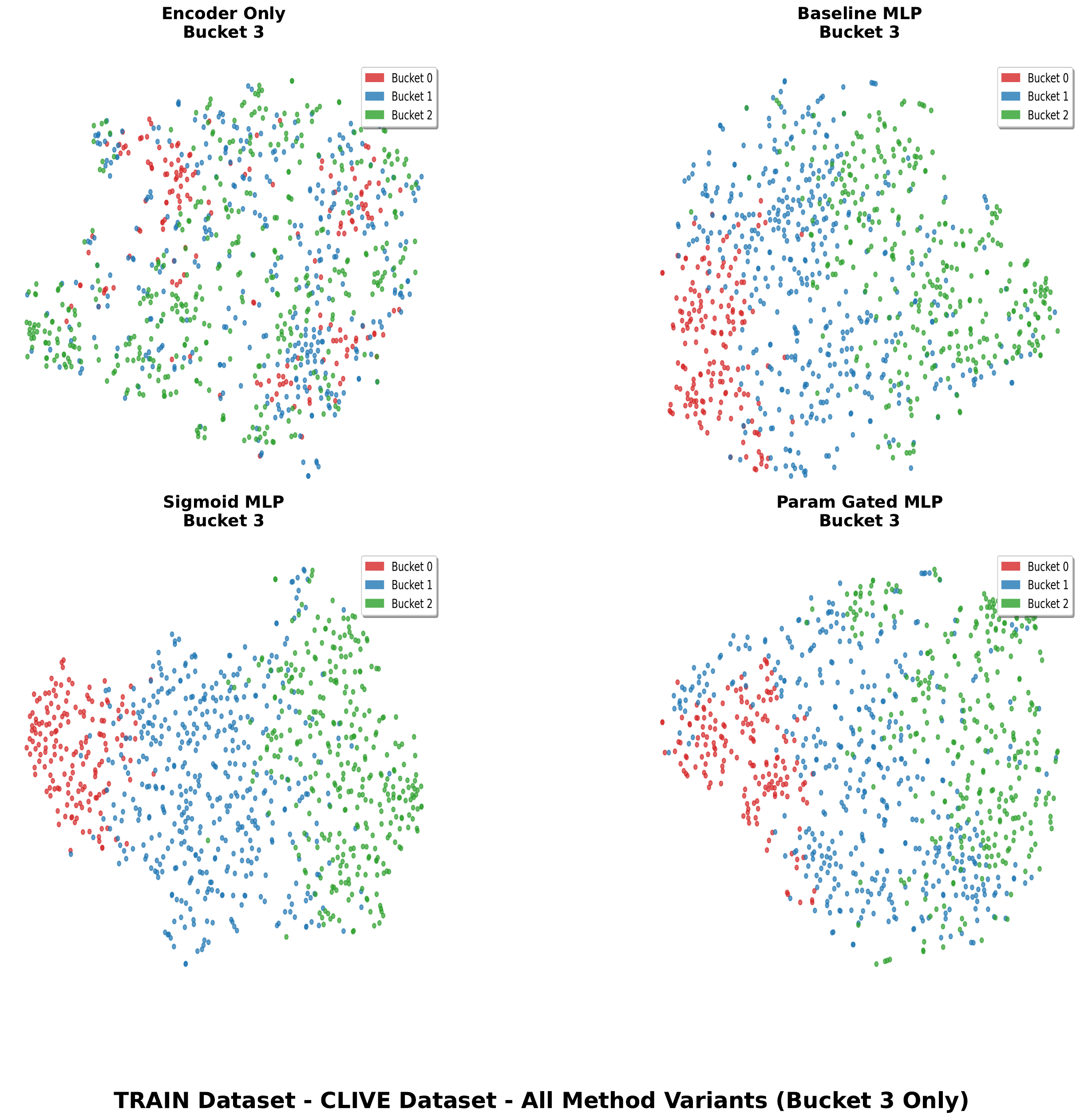}
  \caption{t-SNE comparison of held-out test data (left) versus training data (right) on CLIVE Dataset. We analyze feature representation of different configurations: \textit{Encoder Only} $\rightarrow$ \textit{Baseline MLP} $\rightarrow$ \textit{Sigmoid MLP} $\rightarrow$ 
  \textit{Param-Gated MLP} 
  . Progressively tighter clusters and sharper bucket boundaries indicate the contribution of each module.}

  \label{fig:tsne-model-components-ablation-bucket-wise-3-buckets}
\end{figure*}

We ablate activation functions in the MLP head within the LoRA-augmented 
configuration (Sec.~\ref{sec:exp-setup}). Table~\ref{tab:activation_function_comparison} summarizes 
the results across activation choices. Variants with GELU or Tanh perform 
on par with the baseline LeakyReLU~$\rightarrow$~LeakyReLU , 
showing no clear improvement. In contrast, using a Sigmoid as the first 
activation yields notably stronger results. The most effective design is 
\textbf{+Sig}: Sigmoid~$\rightarrow$~LeakyReLU, which consistently improves SRCC across 
all 3 datasets.
We therefore adopt this 
\textbf{+Sig} configuration  
for all subsequent experiments of the \textbf{MLP head ($d \!\rightarrow\! 512 \!\rightarrow\! 512 \!\rightarrow\! 1$)}, where $d$ denotes the feature dimension of 
the backbone encoder.

\lingqiao{More importantly, we find that employing the sigmoid activation function substantially enhances the generalization ability of the learned NR-IQA models. In particular, cross-dataset experiments (Table~\ref{tab:cross_dataset_results}), where models are trained on one dataset and evaluated on others, show that MLP heads with sigmoid activations yield notably larger performance gains compared to other activation functions.}

\paragraph{Why Sigmoid Activation Works?} \lingqiao{ We hypothesize that the sigmoid function could suppress high magnitude features, which are likely to correspond to the objects, attributes learned in the VL foundation model. While semantic information could be useful for NR-IQA, it might stop the model from exploring lower response features and lead to overfitting. Using the sigmoid function could encourage the model to use evidence from more low-response features.}
Thus, enabling the MLP to learn a balanced quality manifold under noisy MOS labels. t-SNE visualizations (Fig.~\ref{fig:tsne-model-components-ablation-bucket-wise-3-buckets}, supplementary Fig.~\textbf{S2}) support this intuition. LeakyReLU tends to produce entangled semantic clusters, whereas Sigmoid aligns embeddings into perceptual buckets with clear ordinal separation.

\begin{figure}[t]  
  \centering
  \includegraphics[width=0.99\linewidth]{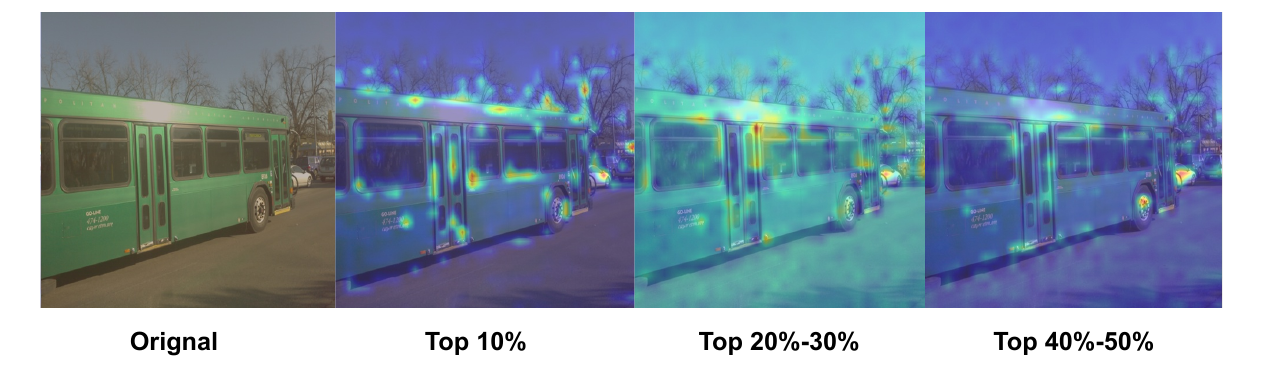}
  \caption{Grad-CAM of SIGLIP2 encoder features, grouped by response magnitude. High-response features align with semantic content, while mid-response features capture subtle artifacts.}
  \label{fig:gradcam_feature_wise}
\end{figure}

Table~\ref{tab:activation_function_comparison} compares four activation functions: Sigmoid, LeakyReLU, ReLU, and GELU. Supplementary Fig.~\textbf{S2} shows that a single Sigmoid gate yields the tightest intra-bucket clusters and clearest inter-bucket margins, translating into an average gain of $\approx$3 SRCC points over LeakyReLU and other activations in low-data regimes \cite{dubey2022activation,zhang2024deep}. This aligns with prior findings that saturating or probabilistic activations are more robust to label noise than piecewise linear units \cite{ven2021regularization}. However, in large datasets ($n \geq 3000$), Sigmoid suffers from vanishing gradients and underperforms LeakyReLU, consistent with the classic limitations of deep sigmoidal networks \cite{he2016deep}. \Ankit{ Table~\ref{tab:sampling_ratio_results_grouped}, Figure~\ref{fig:gradcam_feature_wise}, and Supplementary Figures~S8-S11 further reinforce this limitation. This motivates us to introduce a more flexible non-linearity that the network can adapt during training to optimize performance.}

\paragraph{Gated Activation}

To address this limitation, we introduce an \emph{adaptive} activation blend:
\begin{equation}
\label{eq:gated_mlp}
y = \sigma(g) \odot \,(\gamma\odot\sigma(\alpha \odot x + \beta)) + \bigl(1-\sigma(g)\bigr)\odot\,
      \text{LReLU}_{a}(x)
\end{equation}


where \(\alpha,\beta,\gamma,a,g\) are learnable per channel. This generalizes PReLU \cite{he2015delving} by (i) mixing saturating ~\(\sigma\) and piece-wise-linear LeakyReLU through an adaptive gate and (ii) letting both slopes and offsets vary across channels, a strategy shown to boost expressiveness and convergence.

\textbf{Param Initialization}: We set \(g=0\) so that \(w=\sigma(g)=0.5\), giving equal mixing at the start; for the sigmoid branch we use \((\alpha,\beta)=(1,0)\) to keep it in its natural state; we set the scale \(\gamma=2\) (a fixed choice shown to mitigate vanishing gradients via a “scaling trick” \cite{ven2021regularization}) and let it adapt during training; the Leaky slope starts at \(a=0.25\).
Our gated MLP retains the low-data gains of a pure Sigmoid while matching or surpassing LeakyReLU on large datasets (Table~\ref{tab:gated_mlp_results}). These results support recent evidence that channel-specific, learnable activations offer an effective trade-off between stability and expressive power \cite{hu2018squeeze,sutfeld2020adaptive}. The Gated Activation block is depicted in Figure~\ref{fig:MLP-3-layer}.

Our proposed Gated Activation surpasses prior SOTAs, including diffusion-based approaches, across both transfer directions, with the sole exception of FLIVE $\rightarrow$ CLIVE, demonstrating stronger domain generalization (Table~\ref{tab:cross_dataset_results}).
Figure~\ref{fig:tsne-model-components-ablation-bucket-wise-3-buckets} shows that, in low-data settings, the Gated Activation recovers the same well-separated quality manifold as the single-Sigmoid variant. More importantly, Table~\ref{tab:gated_mlp_results} and Table~\ref{tab:different_backbone_performance_comparison} together demonstrate that the gated design preserves this advantage even in large-data regimes, where the pure Sigmoid head degrades. These results indicate that the gated design adaptively balances activations to match the data distribution and performs across both low- and high-data settings as shown in Figure~\ref{fig: Channel Wise distribution} and supplementary Figure~S7.
\paragraph{Discussion}
\label{sec:gated mlp analysis}
The grouped analysis in Table~\ref{tab:sampling_ratio_results_grouped} shows that Sigmoid-headed MLPs are particularly robust on natural datasets (CLIVE, KonIQ10K), but less so on synthetic ones (AGIQA1K, KADID10K). Natural degradations come from real capture pipelines and are often subtle, meaning they depend on mid-level cues such as texture, noise, and local incoherences that do not strongly correlate with semantic activations Figure~\ref{fig:gradcam_feature_wise}. Synthetic distortions, by contrast, are algorithmic or model-induced and tend to align with prominent edges and structures emphasized by the backbone, reducing the relative advantage of Sigmoid (Supplementary Figures~S8–S11).

Mechanistically, a Sigmoid head compresses the dynamic range and caps large responses, reducing the dominance of strong, semantics-driven features; in contrast, LeakyReLU, being piecewise linear in the positive regime, preserves these large responses. To probe this difference, we design a progressive masking experiment where only the top-$k$ percentile of features ranked by activation strength are retained (Table~\ref{tab:sampling_ratio_results_grouped}). On natural datasets (CLIVE, KonIQ10K), Sigmoid-based MLPs remain stable when high-response features are suppressed but drop sharply when mid-range activations are masked, suggesting that perceptual quality cues reside in the mid-spectrum rather than the strongest semantic activations. LeakyReLU-based MLPs show the opposite behavior, degrading steadily as high-response features are removed, reflecting their reliance on large activations. On synthetic datasets (AGIQA1K, KADID10K), the pattern reverses: distortions align more closely with edges and structures emphasized by strong activations, giving LeakyReLU an advantage while Sigmoid offers less benefit. This interpretation is consistent with the broader distinction between saturating nonlinearities (e.g., Sigmoid) and non-saturating ones (e.g., ReLU family), and with the established use of Sigmoid gates for feature \emph{recalibration/suppression} in architectures such as Squeeze-and-Excitation.

As a baseline for random perturbation, we also compare against dropout 
(Supplementary Table~S3). 
\textit{Input dropout}, which randomly removes embedding dimensions, and \textit{layer dropout}, which randomly perturbs activations after each hidden layer, affect features \emph{uniformly at random} rather than selectively by magnitude. Consequently, they fail to reproduce the structured effects observed under magnitude-based masking, with minimal impact on performance, confirming that our masking results reflect the loci of quality cues in representation space rather than generic regularization. 

\paragraph{Conclusion} Taken together, the evidence suggests that Sigmoid heads act as soft feature suppressors, biasing learning toward mid-response cues that are especially informative for in-the-wild quality prediction, while LeakyReLU heads retain emphasis on large responses, which can be advantageous when quality signals co-vary with salient structure, as in synthetic settings. Grad-CAM visualizations in Figure~\ref{fig:sigmoid_vs_L_RELU_qualitative} and (supplementary) Figure~S6 further provide qualitative support for this interpretation.

\section{Results}


\label{sec:results}

Across seven benchmarks, SigLIP2 is the strongest backbone overall see Tables~\ref{tab:different_backbone_performance_comparison} and \ref{tab:gated_mlp_results}, followed by CLIP and DINOv3, suggesting an advantage of contrastive pre-training over purely self-supervised encoders for NR-IQA. Replacing the first LeakyReLU with a Sigmoid generally improves performance across backbones, with the exception of KADID10K (synthetic distortions), consistent with our observation that Sigmoid activation is most helpful for subtler, natural cues (Section~\ref{sec: Activation function Matters}). Our channel-wise adaptive activation further boosts accuracy beyond both the baseline and Sigmoid variants (Table~\ref{tab:gated_mlp_results}), and unlike the Sigmoid variant, maintains strong performance on both larger and small datasets, indicating better robustness to data scale.

Within non-diffusion methods, our MLP variants outperform prior work, with the adaptive activation leading (Table~\ref{tab:main_results}a). 


In the diffusion-based method comparison (Table~\ref{tab:main_results}b), LGDM is best overall with multi-step refinement, yet our single-step inference closely matches on most benchmarks and outperforms it in cross-dataset transfer (Table~\ref{tab:cross_dataset_results}). Qualitatively, t-SNEs (Figure~\ref{fig:tsne-model-components-ablation-bucket-wise-3-buckets}) corroborate these trends: encoder-only baselines exhibit mixed quality clusters while the Sigmoid head yields clear bucket separation; 
The adaptive variant shows the same pattern on both train and test splits \textit{(see also Supplementary Figures~S1 and S2 for additional t\mbox{-}SNEs)}. 

\section{Conclusion}



Our study establishes a strong, resource‑efficient baseline for no‑reference IQA built on the SigLIP2 foundation.  
First, the vanilla three‑layer head already surpasses most non‑diffusion SOTA methods, confirming the importance of choosing the image encoder of vision–language backbone with rich contrastive pre‑training.  
Second, replacing the first LeakyReLU with a Sigmoid activation yields a consistent performance lift to~+0.034~SRCC on the small data sets like CLIVE, demonstrating that activation selection remains an under‑explored lever in perceptual regression.  
Third, gated activation improves the adaptability to both large and low data regimes 
and also demonstrating near SOTA cross-data performance(see Table~\ref{tab:cross_dataset_results}), highlighting its robustness to domain shift.

\noindent\textbf{Future work}  
We plan to explore (i) richer strategies for mixing nonlinearities,
(ii)~Leveraging the well‑separated embeddings from sigmoid activation, we will investigate knowledge‑distillation schemes to compress the model into lighter backbones for mobile deployment.
{
    \small
    \bibliographystyle{ieeenat_fullname}
    \bibliography{main}
}


\clearpage
\appendix      
\renewcommand{\thefigure}{S\arabic{figure}}
\renewcommand{\thetable}{S\arabic{table}}
\setcounter{figure}{0}
\setcounter{table}{0}

\section{Extended Finetuning and Activation-Function Study}
\subsection{Finetuning Discussion}
\begin{figure*}[t]  
  \centering
  \includegraphics[width=.78\linewidth]{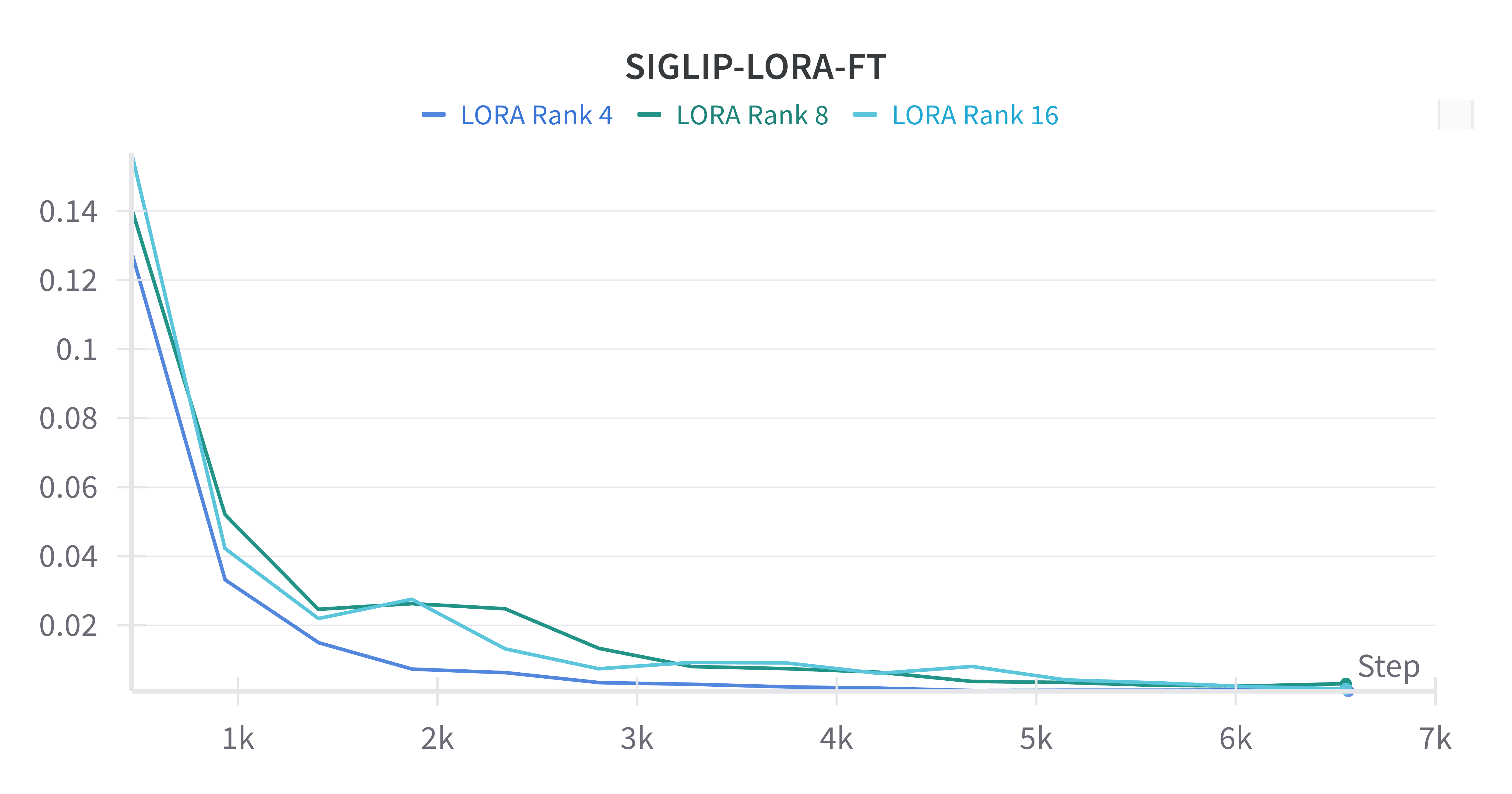}
  \caption{LORA Rank Ablation}
  \label{fig: Lora Rank Ablation}
\end{figure*}

We fix the LoRA rank to 4 in all experiments based on our ablation study (Fig.~\ref{fig: Lora Rank Ablation}) that indicates that rank-4 adapters converge noticeably faster than both lower and higher ranks and achieve this efficiency with minimal memory and compute overhead. For the ResNet encoder, we attach LoRA adapters to the convolutional layers inside each residual block. We choose these layers because they govern spatial filtering and channel mixing, giving high leverage per parameter.

Table \ref{tab:backbone_lora_finetuning_ablation} contrasts two regimes: (i) our full setup with rank-4 LoRA adapters injected into the query and key projections, and (ii) a variant that keeps the visual backbone frozen while training only the lightweight MLP regressor. LoRA fine-tuning yields consistent gains in SRCC/PLCC across all MLP designs and on six of the seven benchmarks. The single exception is FLIVE with the first activation of the MLP head swapped with sigmoid, where the frozen-backbone variant performs slightly better than the LoRA variant. We suspect that the sigmoid activation is the bottleneck that saturates on its 39 K-image scale, capping the head’s capacity. Saturation suppresses gradient flow, so the LoRA adapters cannot harvest their usual gains. We also assess full fine-tuning with both activation heads (baseline and sigmoid). The sigmoid configuration yields slight improvements relative to LoRA, but given LoRA’s markedly better efficiency, we standardize on LoRA for subsequent experiments.

\paragraph{Full Finetuning FT}
For the full finetuning experiments, we keep all other settings identical to the LoRA finetuning configuration but reduce the backbone learning rate to $5\times 10^{-6}$ and set the MLP head learning rate to $1\times 10^{-4}$. This conservative schedule mitigates degradation of the pretrained encoder representations and makes the experiments comparable.

\paragraph{Embedding extraction.}
Unless otherwise stated, we follow the default Hugging Face (HF) implementations and use the encoder’s pooled representation exposed by the model’s forward pass. 
\textit{CLIP/SigLIP:} we call the vision tower and take the projected image embedding (\texttt{image\_embeds}), i.e., pooled visual features passed through the model’s projection head. 
\textit{DINOv2/DINOv3:} we average the last hidden-state tokens (global mean over patch tokens) to obtain a single image vector. 
\textit{Perception Encoder (ViT-L/14-336):} we use the pooled output provided by the HF checkpoint, followed by its projection layer when available. 
\textit{ResNet-152:} we take the \texttt{pooler\_output}, i.e., the global-average-pooled convolutional features returned by \texttt{ResNetModel}. For Diffusion Backbone, We use CleanDIFT\cite{stracke2025cleandift} checkpoints and adopt the DP-IQA\cite{fu2024dp} feature-adapter recipe to aggregate UNet diffusion features into a fixed-length image embedding. All embeddings are then fed to the same prediction head.

\subsection{Activation Function Analysis}




We systematically evaluated all pairwise combinations of four common nonlinearities, Sigmoid, Leaky ReLU, GELU, and Tanh, in the two hidden layers of our three-layer MLP (Table 6, Main Paper). The Sigmoid → Leaky ReLU sequence yields the highest average SRCC and PLCC across the seven NR-IQA benchmarks, while the Sigmoid → Sigmoid variant performs marginally better on a few datasets. Because stacked sigmoids are prone to vanishing gradients, especially under high-data regimes. Hence, we opt for the more stable Sigmoid → Leaky ReLU configuration.

\begin{figure*}[t]  
  \centering
  \includegraphics[width=.48\linewidth]{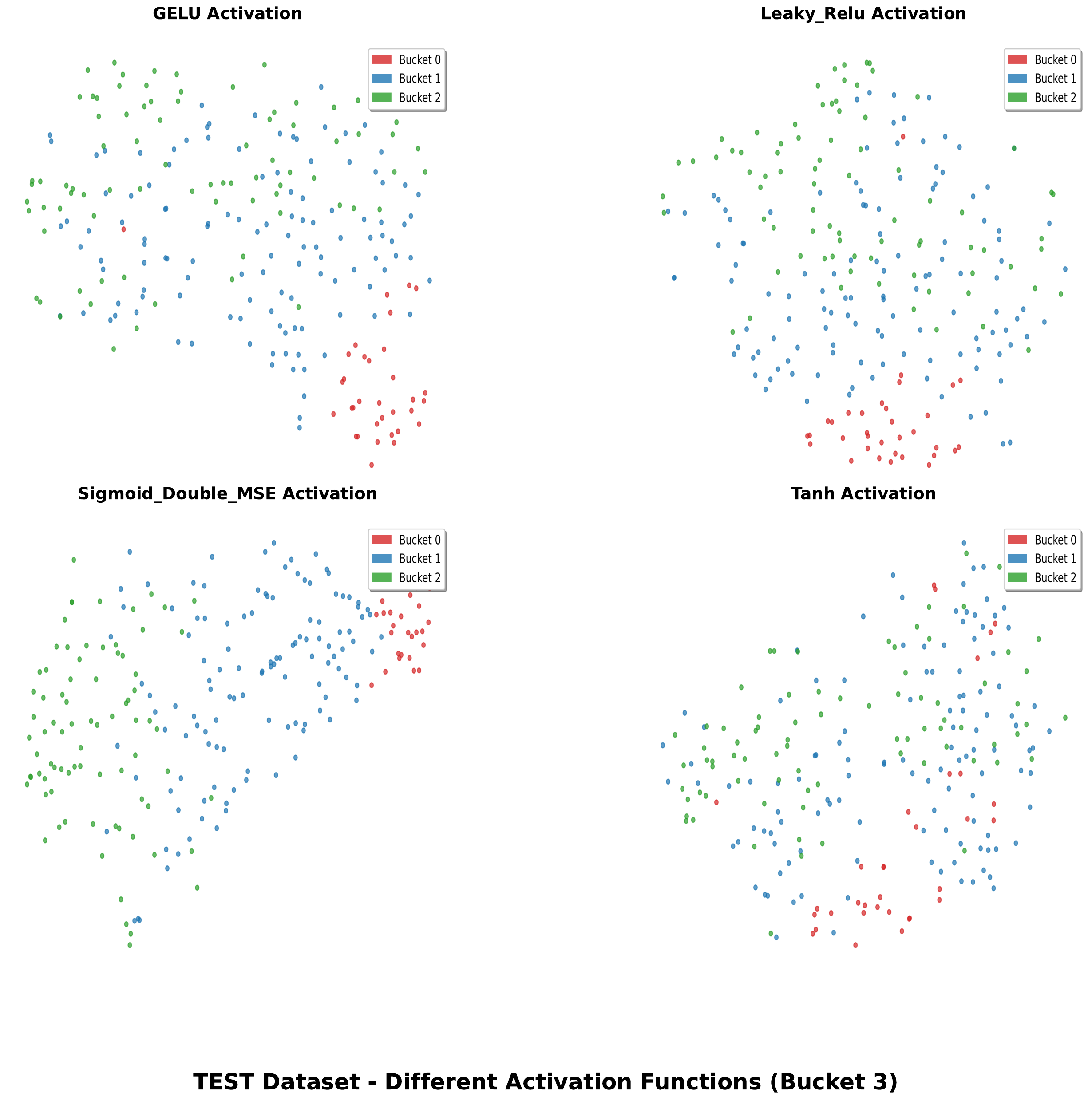}\hfill
  \includegraphics[width=.48\linewidth]{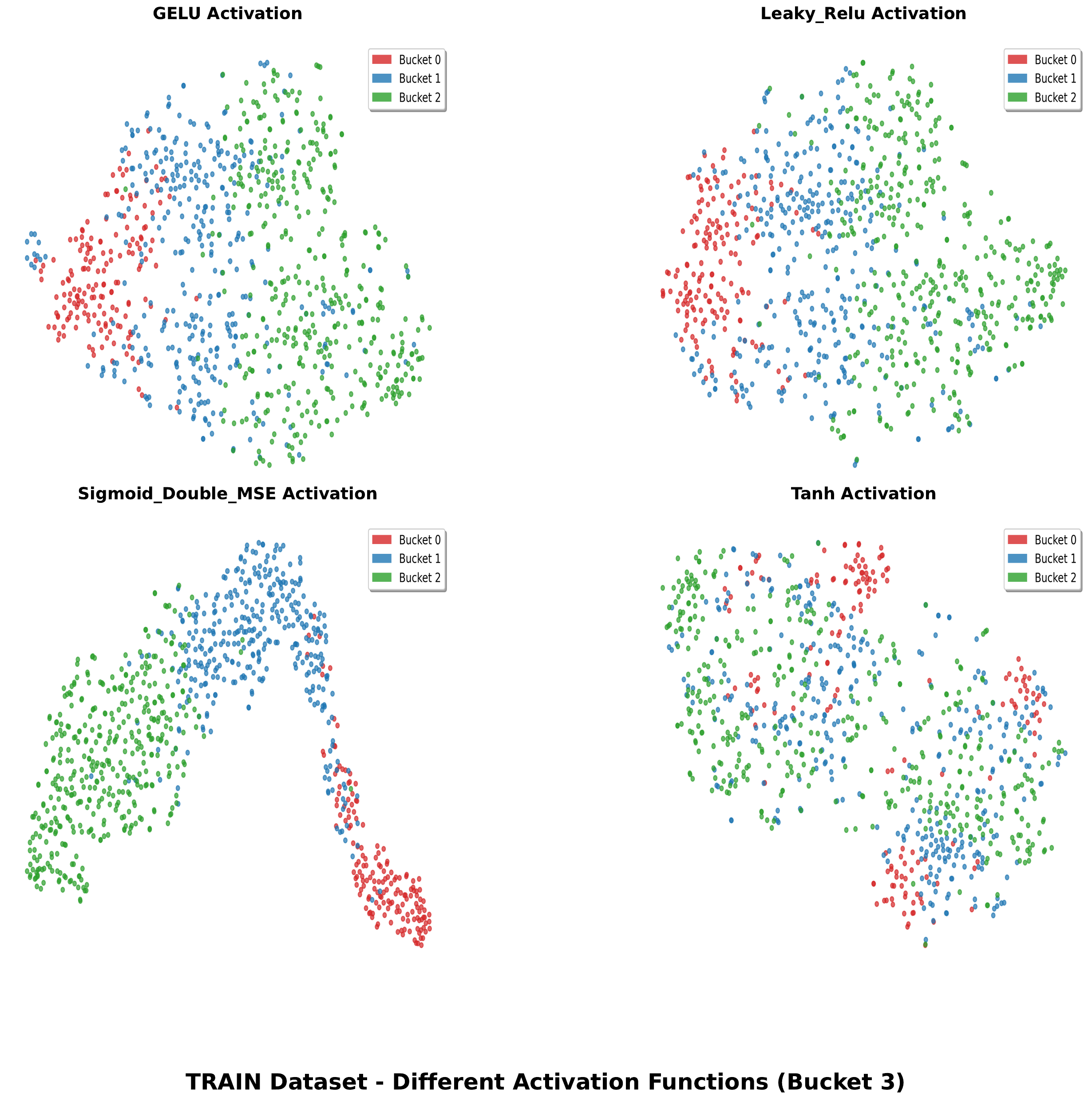}
  \caption{t‑SNE visualizations illustrating the different activation functions.  The left panel depicts embeddings from the held‑out test split, while the right panel shows the corresponding train‑split embeddings.  We observe an interesting phenomenon that Sigmoid activation learns a better representation of the feature space, achieving a better separation.}
  \label{fig: tsne-activation function comparison}
\end{figure*}

GELU offers no statistically significant advantage over Leaky ReLU yet incurs a higher computational cost due to its Gaussian error function; we therefore retain Leaky ReLU as the default second-layer gate. Tanh lags behind all other activations, a trend that is visually corroborated by the fragmented class clusters in the t-SNE embedding of Figure \ref{fig: tsne-activation function comparison}.

We also tested the reverse ordering (Leaky ReLU → Sigmoid, results omitted for brevity); this arrangement negates the convergence benefits provided by the initial sigmoid and does not improve final SRCC Scores. Future work will extend this analysis to recently proposed activations such as Swish \cite{ramachandran2017swish} and Mish \cite{misra2019mish} to further probe their effect on quality-aware feature learning.
\subsection{Learning-Rate Scheduling Strategy for Medium-Scale Datasets}

For all medium-scale datasets (KonIQ-10K, KADID-10K, FLIVE, and SPAQ), we apply a MultiStepLR schedule that lowers the learning rate by a factor of 0.2 at epochs 15 and 25. 

\begin{table*}[!htbp]
\centering
\caption{Ablation study comparing frozen backbones and full fine-tuning (FT) against LoRA fine-tuning on the SigLIP2 backbone. The table shows performance gains from allowing backbone adaptation through LoRA (rank=4) versus keeping the backbone frozen. 
All metrics use "higher is better" scoring. Results are averaged over three runs (seeds: 8, 19, and 25). \textbf{Bold} values indicate the better approach between frozen and LoRA for each configuration. Baseline is a three-layer MLP with LReLU activations.}
\label{tab:backbone_lora_finetuning_ablation}
\footnotesize 
\setlength{\tabcolsep}{3pt} 
\begin{tabular}{@{}l*{14}{c}|*{2}{c}@{}} 
\toprule
\multirow{2}{*}{Method} & \multicolumn{2}{c}{FLIVE} & \multicolumn{2}{c}{SPAQ} & \multicolumn{2}{c}{CLIVE} & \multicolumn{2}{c}{AGIQA3K} & \multicolumn{2}{c}{KADID10K} & \multicolumn{2}{c}{KonIQ10K} & \multicolumn{2}{c|}{AGIQA1K} & \multicolumn{2}{c}{Average} \\
\cmidrule(lr){2-3} \cmidrule(lr){4-5} \cmidrule(lr){6-7} \cmidrule(lr){8-9} \cmidrule(lr){10-11} \cmidrule(lr){12-13} \cmidrule(lr){14-15} \cmidrule(lr){16-17}
& SRCC & PLCC & SRCC & PLCC & SRCC & PLCC & SRCC & PLCC & SRCC & PLCC & SRCC & PLCC & SRCC & PLCC & SRCC & PLCC \\
\midrule
Baseline (Frozen) & .451 & .544 & .902 & .905 & .822 & .841 & .858 & .913 & .944 & .945 & .874 & .895 & .853 & \textbf{.891} & .786 & .818 \\
Baseline (FT) & .509 & .630 & .885 & .890 & .730 & .742 & .632 & .723 & .685 & .689 & .756 & .781 & .779 & .837 & .711 & .756 \\
Baseline (LoRA) & \textbf{.533}& \textbf{.641}& \textbf{.927}& \textbf{.931}& \textbf{.875}& \textbf{.905}& \textbf{.865}& \textbf{.917}& \textbf{.961}& \textbf{.964}& \textbf{.932}& \textbf{.943}& \textbf{.857}& .889& \textbf{.844}& \textbf{.879}\\
\midrule
Baseline (Frozen)+Sig & .537 & .620 & .896 & .899 & .827 & .852 & .861 & .916 & .913 & .911 & 0.888 & 0.906 & .842 & .880 & .818 & .850 \\
Baseline (FT)+Sig & \textbf{.561} & \textbf{.659} & \textbf{.923} & \textbf{.928} & .882 & .908 & .864 & .918 & \textbf{.963} & \textbf{.966} & .931 & .946 & .845 & .880 & \textbf{.853} & \textbf{.886} \\
Baseline (LoRA)+Sig & .521& .608& .921& .926& \textbf{.909}& \textbf{.930}& \textbf{.878}& \textbf{.923}& .939& .943& \textbf{.938}& \textbf{.947}& \textbf{.872}& \textbf{.897}& .846& .874\\
\bottomrule
\end{tabular}
\end{table*}



\section{Experimental Variance Analysis}
\label{sec: Experimental Variance Analysis}

Each configuration is trained and evaluated three times with independent random seeds (8, 19, 25). We report the seed-averaged results in the main tables and list the associated standard deviations in Table \ref{tab:master_std} and Table~\ref{tab:cross_std} for cross-dataset experiments. 

\begin{figure*}[t]  
  \centering
  \includegraphics[width=.48\linewidth]{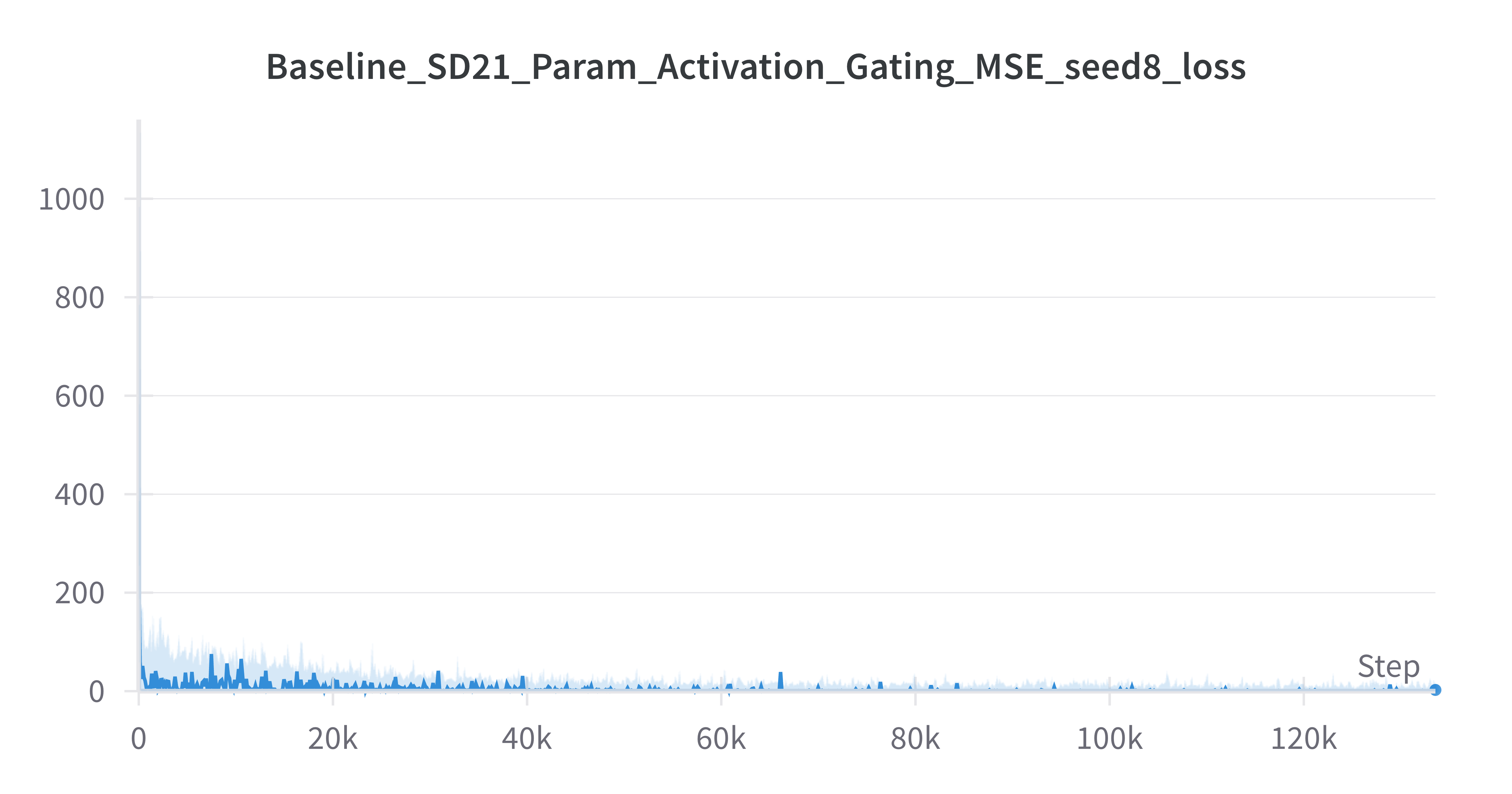}\hfill
  \includegraphics[width=.48\linewidth]{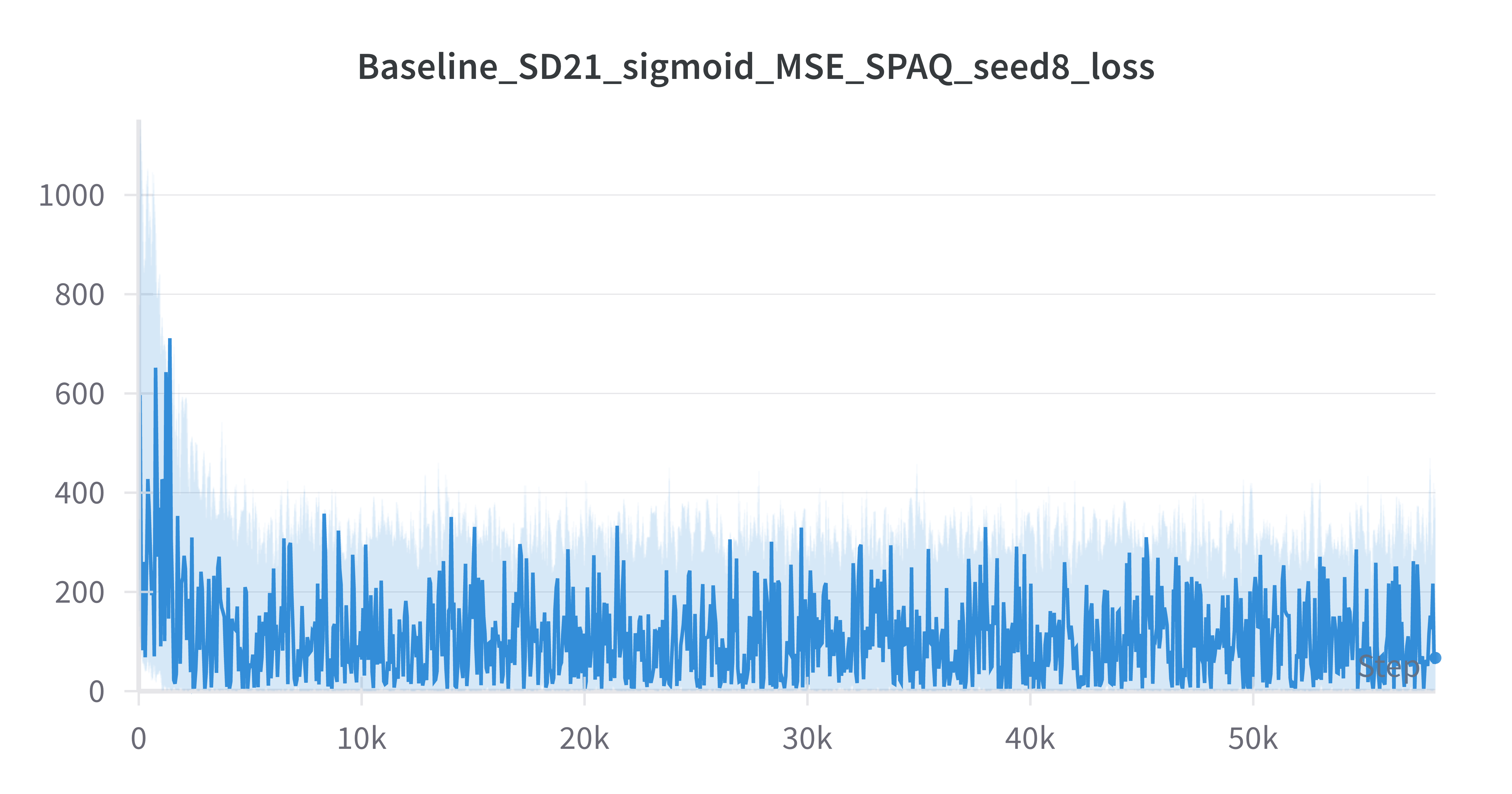}
  \caption{Training loss on SPAQ with the CleanDIFT–SD 2.1 backbone. \textbf{Left:} MLP head whose first two hidden layers use LeakyReLU, showing smooth convergence. \textbf{Right:} identical MLP but with a Sigmoid in the first hidden layer; the loss spikes and remains unstable, illustrating the saturation-induced optimisation failure discussed in Section \ref{sec: Experimental Variance Analysis}.}
  \label{fig: Loss Curve for SPAQ}
\end{figure*}

During these trials, we encountered sporadic numerical instabilities when pairing the CleanDIFT-based SD2.1 backbone with the Sigmoid-first MLP on the SPAQ dataset. In Figure~\ref{fig: Loss Curve for SPAQ} we can notice that the loss explodes for the experiment with the Sigmoid activation in the first hidden layer as it steepens the input distribution, causing many neurons to saturate, the resulting near-zero gradients prevent effective weight updates and precipitate optimisation instability, which is evident in the Figure~\ref{fig: Loss Curve for SPAQ} where the loss remains unchanged. On large-scale datasets such as SPAQ with 40 K samples, the higher number of parameter updates amplifies this vanishing-gradient problem, leading to persistent, high-variance losses and eventual training collapse. We hypothesize that the consistent drop in performance of Sigmoid MLP on larger datasets is due to the same effect; hence, introducing a parallel LeakyReLU branch (our gated activation) restores non-zero gradients, thereby stabilizing training across both small and large data regimes.

\subsection{Training Setting}
All experiments are run in mixed precision. Model weights are stored in FP16, while the parameters of the learnable activation gates remain in full FP32 to preserve numerical range and prevent gradient underflow. This hybrid setting maintains the speed and memory benefits of half-precision training without compromising the convergence of the gated activations. We train with a physical batch size of~2 and gradient‑accumulation of~6, yielding an effective batch size of~12. All experiments were executed on a single NVIDIA A100 GPU. Performance metrics are reported in Table~\ref{tab:compute_siglip2}

\begin{figure*}[t]  
  \centering
  \includegraphics[width=.78\linewidth]{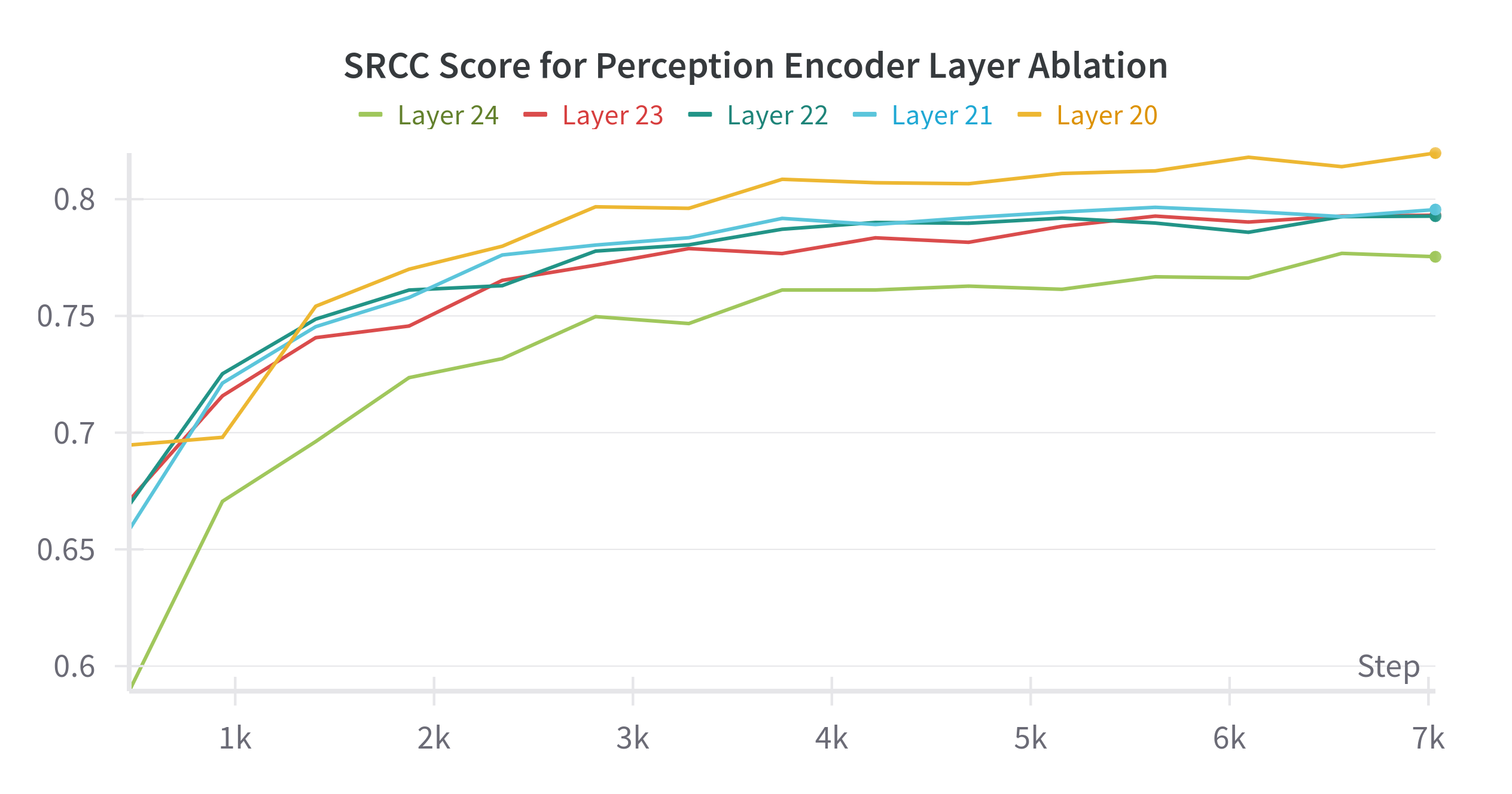}
  \caption{Perception Encoder Layers Ablation (CLIVE Dataset)}
  \label{fig: Perception Encoder Layers Ablation}
\end{figure*}

\paragraph{Perception Backbone}
Building on the observation by Bolya \textit{et al.}\citep{bolya2025perception} that intermediate Perception features can boost downstream performance, we conducted a layer-selection sweep for the \textbf{ViT-L/14-336} checkpoint, an ablation not reported in the original paper (see Figure \ref{fig: Perception Encoder Layers Ablation}). Empirically, features tapped at layer 20 yield the highest validation SRCC on the NR-IQA task, outperforming neighboring layers. Accordingly, all subsequent experiments that use the Perception backbone extract features from layer 20, ensuring a fair and capacity-maximizing comparison with the other encoders.

\paragraph{Dataset Split}
For every dataset in this study, we perform an 80 / 20 random split training versus validation using the three seed values specified above. The identical protocol is applied to all datasets to ensure consistent evaluation and fair cross-dataset comparisons. 


\paragraph{Normalization}
Unless otherwise noted, we map all opinion scores to the range \([0,1]\) for a uniform training target. 
Concretely, we rescale KonIQ-10k MOS by \(y/5\) (official MOS are on a 5-point ACR scale), 
SPAQ MOS by \(y/100\) (scores reported on a 0–100 scale), 
CLIVE MOS by \(y/100\) (LIVE Challenge database), 
FLIVE MOS by \(y/100\), 
AGIQA-3K by \(y_{\text{quality}}/5\) and \(y_{\text{align}}/5\) (the release provides normalized MOS columns; we standardize to \([0,1]\) regardless), 
AGIQA-1K by \(y/5\) (normalized MOS in the official spreadsheet), 
and KADID-10k DMOS by \((y-1)/4\) to convert the \([1,5]\) range to \([0,1]\) with higher being better. 

\section{Embedding Response Analysis}

\begin{figure*}[t]  
  \centering
  \includegraphics[width=.78\linewidth]{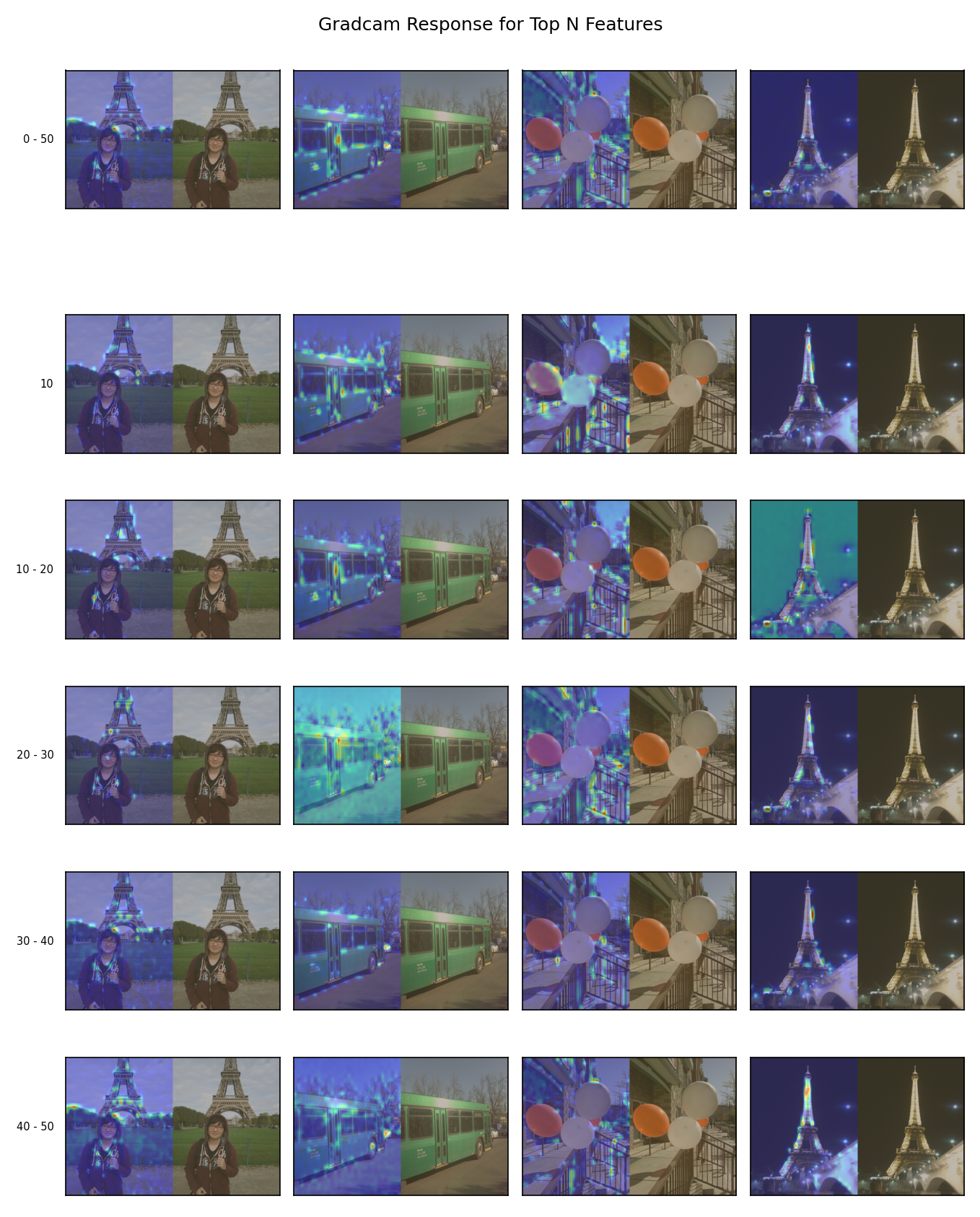}
  \caption{Comparison of Grad-CAM visualizations across SIGLIP2 encoder feature groups on CLIVE images, showing the correspondence between feature responses and input regions. Here, Top-N refers to features whose absolute magnitudes are at or above the Nth percentile.}
  \label{fig: Grad-CAM feature CLIVE}
\end{figure*}
\begin{figure*}[t]  
  \centering
  \includegraphics[width=.78\linewidth]{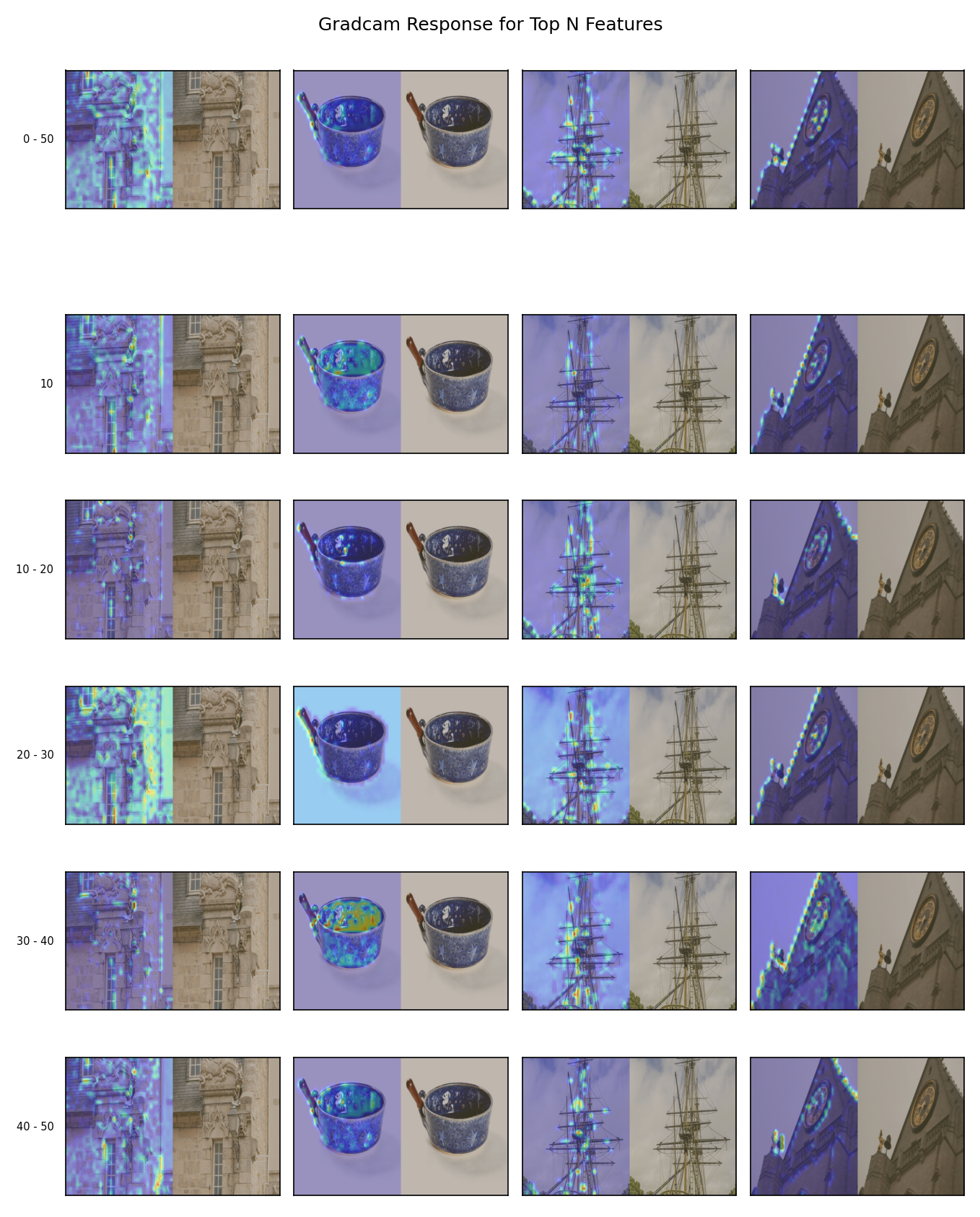}
  \caption{Comparison of Grad-CAM visualizations across SIGLIP2 encoder feature groups on KonIQ-10K images, showing the correspondence between feature responses and input regions. Here, Top-N refers to features whose absolute magnitudes are at or above the Nth percentile.}
  \label{fig: Grad-CAM feature KonIQ10K}
\end{figure*}
\begin{figure*}[t]  
  \centering
  \includegraphics[width=.78\linewidth]{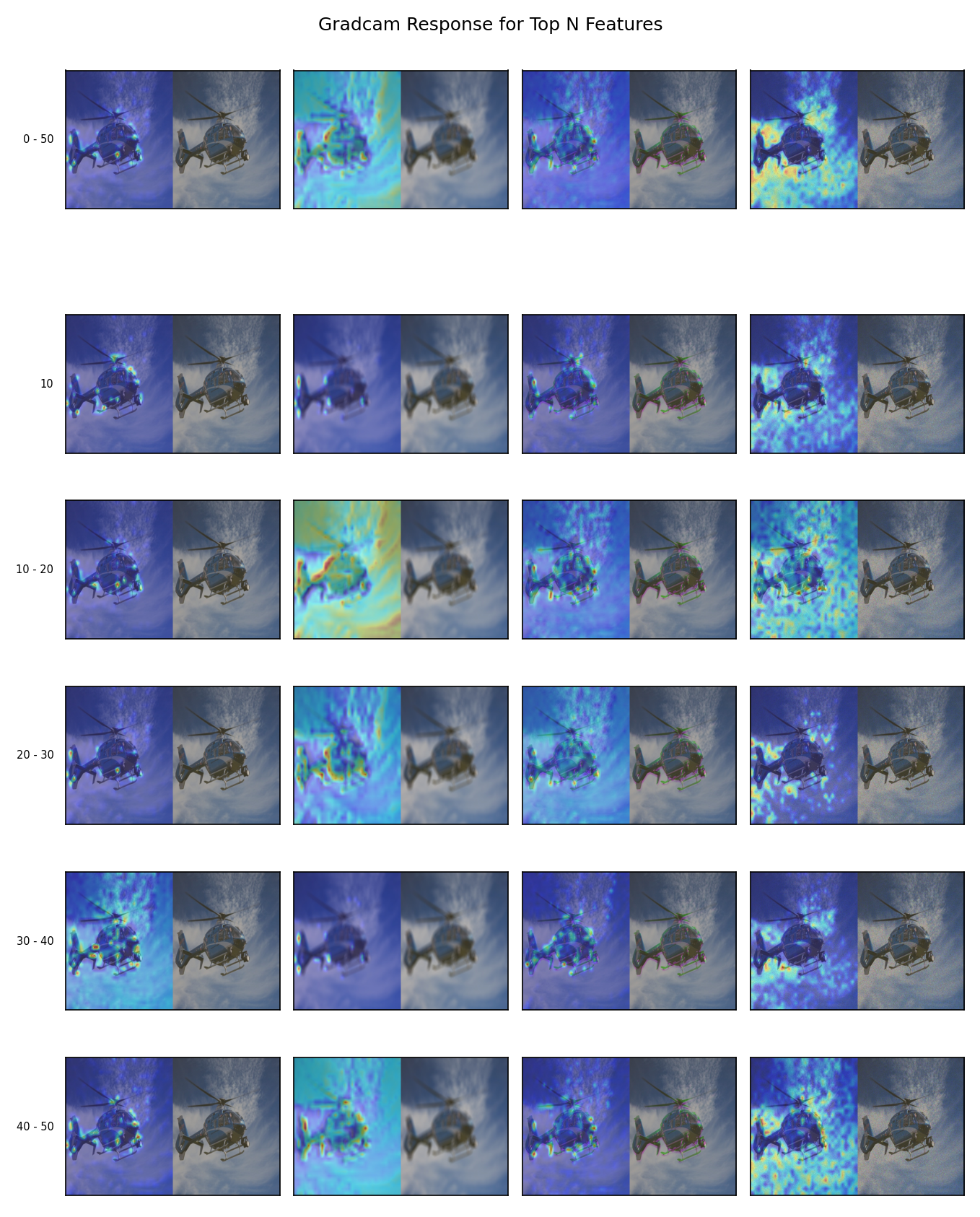}
  \caption{Comparison of Grad-CAM visualizations across SIGLIP2 encoder feature groups on KADID-10K images, showing the correspondence between feature responses and input regions. Here, Top-N refers to features whose absolute magnitudes are at or above the Nth percentile.}
  \label{fig: Grad-CAM feature KADID10K}
\end{figure*}
\begin{figure*}[t]  
  \centering
  \includegraphics[width=.78\linewidth]{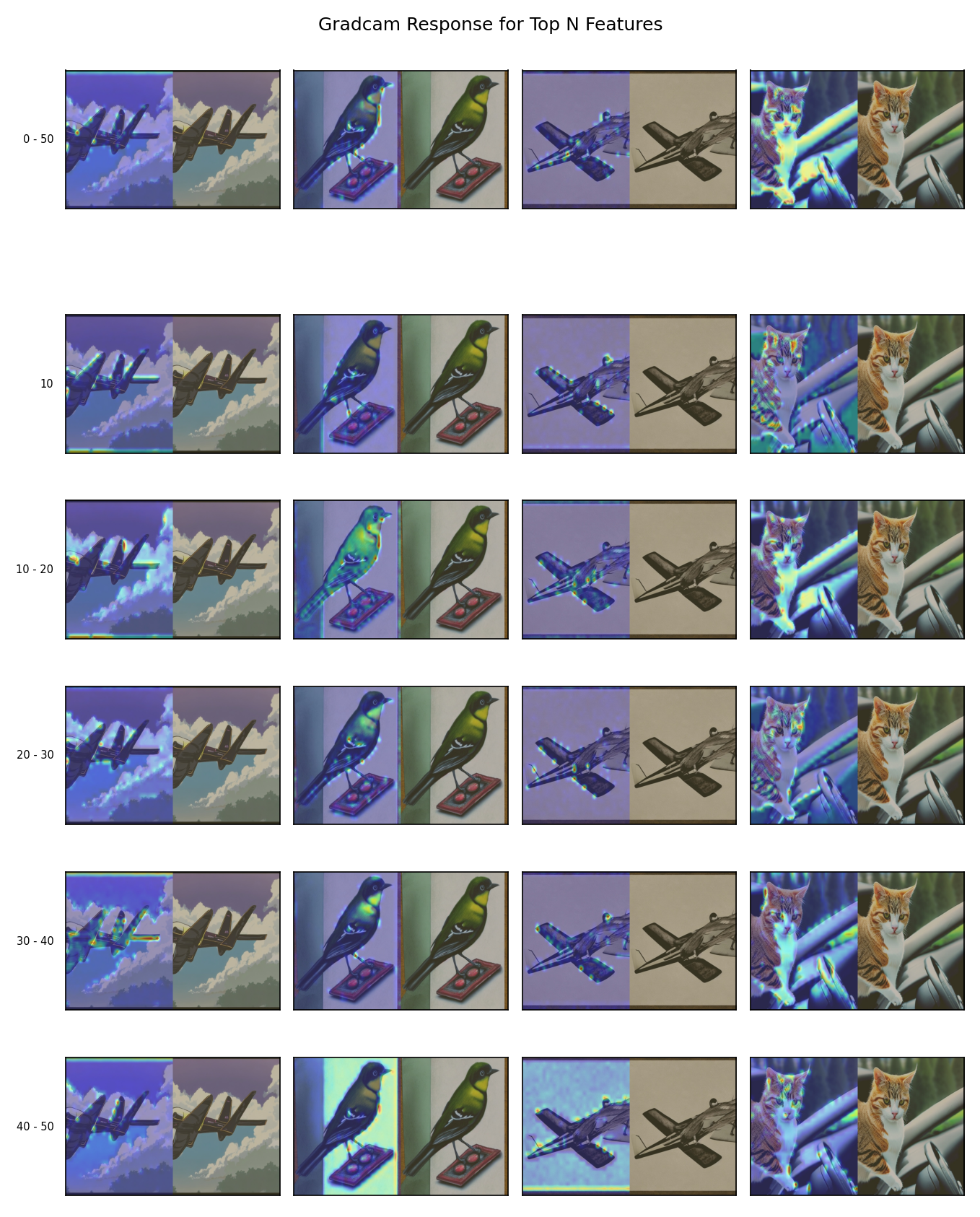}
  \caption{Comparison of Grad-CAM visualizations across SIGLIP2 encoder feature groups on AGIQA-1K images, showing the correspondence between feature responses and input regions. Here, Top-N refers to features whose absolute magnitudes are at or above the Nth percentile.}
  \label{fig: Grad-CAM feature AGIQA1K}
\end{figure*}

\paragraph{Experiment}
We use a pretrained SigLIP2 encoder to compute image embeddings and rank feature activations by absolute magnitude. For each percentile band \(P\), we retain features whose magnitudes fall within \(P\) and generate Grad-CAM heatmaps on the original image from those features. We visualize four randomly sampled images from each of four datasets, CLIVE, KonIQ10K, KADID10K, and AGIQA1K. CLIVE and KonIQ10K contain authentic distortions; AGIQA1K comprises AI-generated images; KADID10K applies synthetic distortions to natural images. We visualize Top-\(N\%\) percentile bands (with \(N\in[1,50]\)), letting \(S=\{|f_i|\}\), \(\text{Top-}N\%\!=\!\{\,i:\,|f_i|\ge Q_{1-N/100}(S)\,\}\). See Figures~\ref{fig: Grad-CAM feature CLIVE}–\ref{fig: Grad-CAM feature AGIQA1K}.

\paragraph{Observations}
In our feature attribution analyses, we consistently find that the highest-ranked channels (top decile by response magnitude) correlate most strongly with semantic structure, whereas mid-ranked channels capture more general contextual regularities. The precise band that encodes this “mid-level” context varies across images (e.g., ranks 20–30 for some scenes and 30–40 for others), but the pattern persists. The very top responses align with object and layout level semantics. This helps explain the gains we observe with a sigmoid first-layer activation, by compressing extremes and enlarging sensitivity in the mid-range, sigmoid implicitly regularizes the head to exploit these mid-level features, improving robustness and generalization (Table 4, main paper). Dataset behavior further supports this interpretation. On AGIQA-1K, where degradations are primarily generative artifacts that disrupt global semantics (e.g., ill-formed content and structural inconsistencies), quality is tightly coupled to semantic fidelity. Similarly, KADID-10k comprises controlled, intensity-graded distortions; several of these are well captured by strong low-level departures in the feature space, for which LeakyReLU’s near-linear pass-through at large magnitudes remains advantageous, consistent with its competitive results on this benchmark. By contrast, on natural-image datasets like CLIVE and KonIQ-10K, where quality information lies in subtle perceptual cues, the sigmoid head excels by prioritizing the mid-to-high semantic regime.

\begin{table*}[!htbp]
\centering
\caption{Standard deviations of SRCC and PLCC across all experiments. Values represent standard deviation across three runs (seeds: 8, 19, 25). This table provides STD values for all experiments mentioned throughout the paper. \textbf{Note:} $^\dagger$ Abnormally high STD values suggest potential instability in these configurations.}
\label{tab:master_std}
\footnotesize
\setlength{\tabcolsep}{2.5pt}
\begin{tabular}{@{}l*{14}{c}@{}}
\toprule
& \multicolumn{2}{c}{\textbf{FLIVE}} & \multicolumn{2}{c}{\textbf{SPAQ}} & \multicolumn{2}{c}{\textbf{CLIVE}} & \multicolumn{2}{c}{\textbf{AGIQA3K}} & \multicolumn{2}{c}{\textbf{KADID10K}} & \multicolumn{2}{c}{\textbf{KonIQ10K}} & \multicolumn{2}{c}{\textbf{AGIQA1K}} \\
\cmidrule(lr){2-3} \cmidrule(lr){4-5} \cmidrule(lr){6-7} \cmidrule(lr){8-9} \cmidrule(lr){10-11} \cmidrule(lr){12-13} \cmidrule(lr){14-15}
\textbf{Experiment} & SRCC & PLCC & SRCC & PLCC & SRCC & PLCC & SRCC & PLCC & SRCC & PLCC & SRCC & PLCC & SRCC & PLCC \\
\midrule
CLIP & .0174 & .0002 & .0019 & .0006 & .0245 & .0014 & .0082 & .0019 & .0014 & .0008 & .0056 & $<$.0001 & .0081 & .0024 \\
CLIP + Sigmoid & .0048 & .0001 & .0033 & .0009 & .0174 & .0020 & .0088 & .0016 & .0021 & .0006 & .0009 & $<$.0001 & .0088 & .0013 \\
CLIP + Activation Gating & .0097 & .0002 & .0017 & .0005 & .0182 & .0016 & .0069 & .0020 & .0009 & .0007 & .0024 & .0001 & .0172 & .0018 \\

DINO & .0051 & .0003 & .0014 & .0007 & .0436 & .0052 & .0042 & .0019 & .0011 & .0008 & .0136 & $<$.0001 & .0112 & .0046 \\
DINO + Sigmoid & .0062 & .0003 & .0007 & .0012 & .0197 & .0060 & .0050 & .0029 & .0082 & .0030 & .0039 & $<$.0001 & .0073 & .0036 \\
DINO + Activation Gating & .0054 & .0002 & .0007 & .0010 & .0223 & .0049 & .0099 & .0006 & .0073 & .0012 & .0023 & .0004 & .0103 & .0024 \\

DINO-3 & .0093 & .0002 & .0005 & .0003 & .0410 & .0089 & .0065 & .0013 & .0008 & .0021 & .0019 & .0005 & .0151 & .0017 \\
DINO-3 + Sigmoid & .0029 & .0002 & .0009 & .0012 & .0247 & .0065 & .0110 & .0016 & .0031 & .0015 & .0050 & .0003 & .0128 & .0013 \\
DINO-3 + Activation Gating & .0014 & .0003 & .0013 & .0005 & .0358 & .0051 & .0078 & .0018 & .0022 & .0005 & .0019 & .0005 & .0138 & .0011 \\

ResNet152 & .0290 & .0022 & .0054 & .0020 & .0483 & .0120 & .0316 & .0013 & .0063 & .0047 & .0072 & .0001 & .0349 & .0042 \\
ResNet152 + Sigmoid & .0429 & .0004 & .0013 & .0010 & .0213 & .0139 & .0334 & .0028 & .0234 & .0079 & .0376 & .0001 & .0090 & .0053 \\
ResNet152 + Activation Gating & .0019 & .0008 & .0037 & .0011 & .0454 & .0068 & .0358 & .0028 & .0039 & .0012 & .0226 & .0013 & .0031 & .0041 \\

Perception & .0067 & .0003 & .0015 & .0004 & .0167 & .0050 & .0084 & .0026 & .0006 & .0003 & .0211 & .0001 & .0222 & .0024 \\
Perception + Sigmoid & .0065 & .0002 & .0005 & .0003 & .0198 & .0052 & .0095 & .0008 & .0014 & .0007 & .0107 & $<$.0001 & .0084 & .0016 \\
Perception + Activation Gating & .0052 & .0001 & .0016 & .0008 & .0150 & .0040 & .0080 & .0004 & .0003 & .0013 & .0031 & .0004 & .0177 & .0030 \\

SigLIP2 (Frozen) & .0076 & $<$.0001 & .0024 & .0004 & .0114 & .0007 & .0093 & .0007 & .0008 & .0009 & .0166 & .0001 & .0119 & .0030 \\
SigLIP2 (Frozen) + Sigmoid & .0037 & .0001 & .0026 & .0004 & .0105 & .0021 & .0120 & .0018 & .0019 & .0016 & .0027 & $<$.0001 & .0135 & .0008 \\

SigLIP2 (FT) & .0065 & .0001 & .0491 & .0099 & .1771 & .0244 & .3099 & .0138 & .3950 & .0398 & .2223 & .0174 & .0977 & .0096 \\
SigLIP2 (FT) + Sigmoid & .0009 & .0003 & .0014 & .0003 & .0099 & .0031 & .0054 & .0017 & .0010 & .0002 & .0039 & .0003 & .0126 & .0015 \\

SigLIP2 & .0416 & .0009 & .0010 & .0006 & .0103 & .0034 & .0116 & .0009 & .0017 & .0007 & .0051 & $<$.0001 & .0080 & .0019 \\
SigLIP2 + Sigmoid & .0047 & .0048 & .0037 & .0006 & .0090 & .0015 & .0092 & .0012 & .0060 & .0015 & .0143 & $<$.0001 & .0055 & .0013 \\
SigLIP2 + Activation Gating & .0224 & .0014 & .0023 & $<$.0001 & .0054 & .0026 & .0092 & .0015 & .0014 & .0007 & .0039 & .0003 & .0054 & .0018 \\

CleanDIFTSD2.1 & --- & 0.013 & .0022 & 3.918$^\dagger$ & .0294 & .0071 & .0097 & .0185 & .0012 & .0003 & .0425 & .0002 & .0059 & .0260 \\
CleanDIFTSD2.1 + Sigmoid & .0007 & .0014 & 0.000 & 43.85$^\dagger$ & .0217 & .0057 & .0121 & .0147 & .0008 & .0006 & .0032 & $<$.0001 & .0062 & .0265 \\
CleanDIFTSD2.1 + Activation Gating & .0040 & .0002 & .0033 & 3.915$^\dagger$ & .0308 & .0066 & .0042 & .0194 & .0009 & .0005 & .0043 & .0003 & .0060 & .0347 \\
\bottomrule
\end{tabular}
\end{table*}

\begin{figure*}[t]  
  \centering
  \includegraphics[width=.48\linewidth]{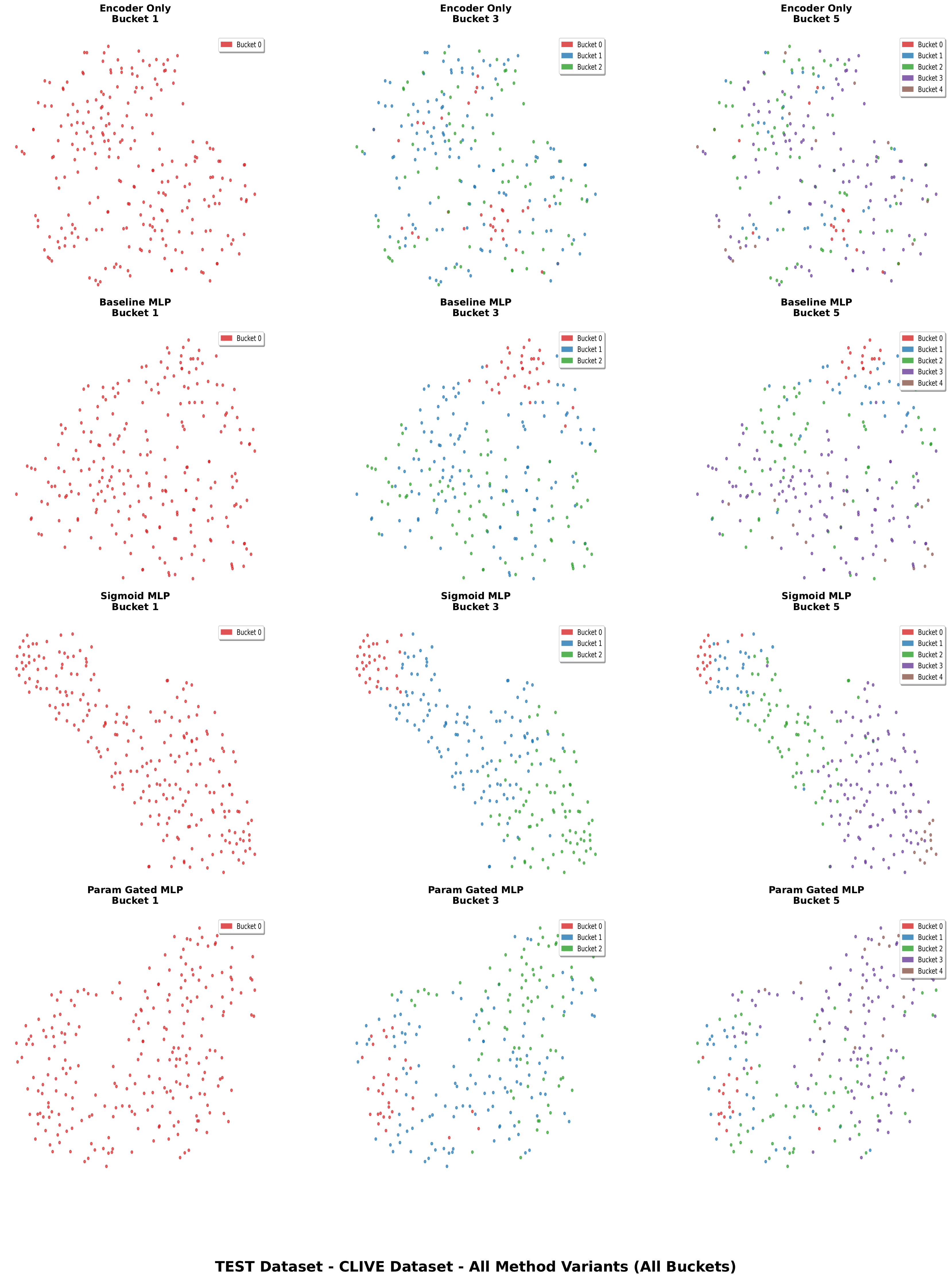}\hfill
  \includegraphics[width=.48\linewidth]{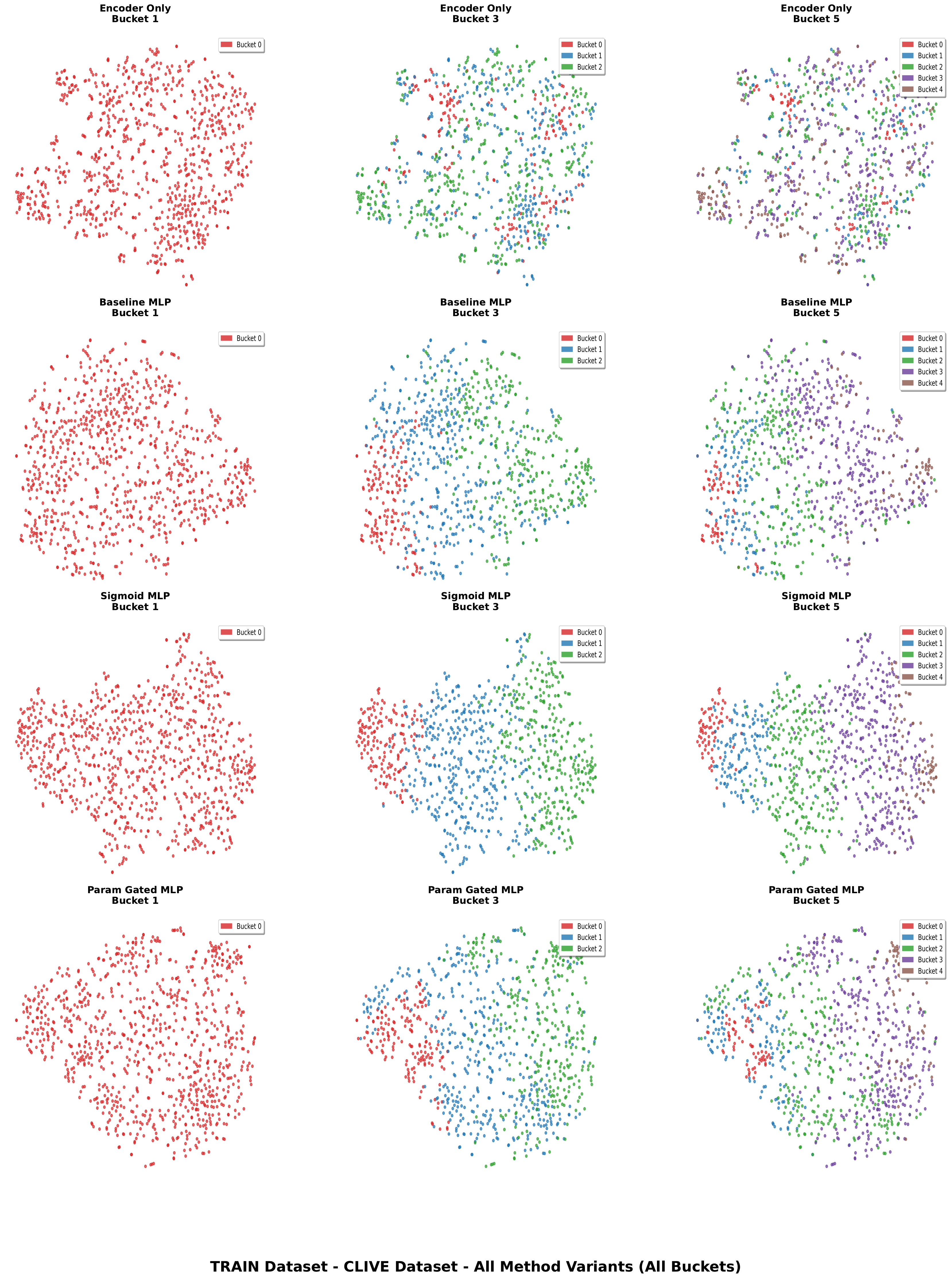}
  \caption{t‑SNE visualizations illustrating the contribution of each architectural component.  The left panel depicts embeddings from the held‑out test split, while the right panel shows the corresponding train‑split embeddings.  Clear separation across quality buckets in the test plot indicates that the learned representation generalizes beyond the training data.}

  \label{fig:tsne-model-components-ablation-bucket wise}
\end{figure*}


\begin{table*}[t]
\centering
\caption{Comparison of MLP performance under different feature-retention percentiles $k$ \emph{(dropout variants)}. Reported are the mean best Spearman (SRCC) and Pearson (PLCC) for Group~1 (synthetic: AGIQA1K, KADID10K) and Group~2 (natural: CLIVE, KonIQ10K). $\Delta$ denotes the change relative to $k=100$ within each group (more negative indicates a larger drop). Here, $k$ is the retained percentile of features by magnitude used in training/evaluation. \emph{For dropout experiments, the dropout rate is $(1-k)$ (i.e., $k$ is the keep probability).}}
\label{tab:sampling_ratio_results_grouped_dropout}
\small
\begin{tabular}{@{}lc*{4}{c}|*{2}{c}|*{4}{c}|*{2}{c}@{}}
\toprule
\multirow{2}{*}{Method} & \multirow{2}{*}{k} 
& \multicolumn{2}{c}{AGIQA1K} & \multicolumn{2}{c}{KADID10K} 
& \multicolumn{2}{|c|}{$\Delta$ Group 1} 
& \multicolumn{2}{c}{CLIVE} & \multicolumn{2}{c}{KonIQ10K} 
& \multicolumn{2}{|c}{$\Delta$ Group 2} \\
\cmidrule(lr){3-4} \cmidrule(lr){5-6} \cmidrule(lr){7-8} \cmidrule(lr){9-10} \cmidrule(lr){11-12} \cmidrule(lr){13-14}
& & SRCC & PLCC & SRCC & PLCC & SRCC & PLCC & SRCC & PLCC & SRCC & PLCC & SRCC & PLCC \\
\midrule

\multirow{6}{*}{\parbox{2.5cm}{MLP-LReLU w/\\Dropout Layers}} 
& 100 & .854 & .890 & .891 & .892 & 0 & 0 & .875 & .906 & .908 & .917 & 0 & 0 \\
& 90  & .855 & .890 & .883 & .884 & -0.004 & -0.004 & .875 & .906 & .915 & .912 & -0.001 & -0.003 \\
& 70  & .857 & .887 & .869 & .871 & -0.009 & -0.011 & .880 & .907 & .896 & .896 & -0.008 & -0.015 \\
& 50  & .850 & .886 & .869 & .853 & -0.013 & -0.021 & .878 & .906 & .889 & .889 & -0.008 & -0.014 \\
& 30  & .826 & .868 & .865 & .861 & -0.027 & -0.027 & .865 & .896 & .876 & .878 & -0.021 & -0.025 \\
& 10  & .738 & .820 & .861 & .838 & -0.073 & -0.062 & .812 & .812 & .864 & .876 & -0.054 & -0.067 \\
\midrule

\multirow{6}{*}{\parbox{2cm}{MLP-LReLU w/\\Input Dropout}} 
& 100 & .855 & .891 & .885 & .888 & 0 & 0 & .877 & .909 & .915 & .918 & 0 & 0 \\
& 90  & .854 & .888 & .881 & .884 & -0.003 & -0.004 & .876 & .905 & .913 & .914 & -0.002 & -0.004 \\
& 70  & .851 & .888 & .868 & .870 & -0.011 & -0.011 & .879 & .907 & .904 & .917 & -0.004 & -0.003 \\
& 50  & .852 & .885 & .864 & .865 & -0.012 & -0.019 & .880 & .910 & .892 & .896 & -0.010 & -0.011 \\
& 30  & .844 & .880 & .858 & .842 & -0.019 & -0.028 & .882 & .909 & .891 & .892 & -0.009 & -0.016 \\
& 10  & .811 & .853 & .760 & .763 & -0.084 & -0.081 & .847 & .872 & .872 & .885 & -0.037 & -0.047 \\
\bottomrule
\end{tabular}
\end{table*}

\begin{table*}[!htbp]
\centering
\caption{Standard deviations of SRCC and PLCC for cross-dataset evaluations (seeds: 8, 19, 25). Values rounded to 4 decimals.}
\label{tab:cross_std}
\footnotesize
\setlength{\tabcolsep}{4pt}
\begin{tabular}{@{}l*{8}{c}@{}}
\toprule
& \multicolumn{2}{c}{\textbf{Pair 1}} & \multicolumn{2}{c}{\textbf{Pair 2}} & \multicolumn{2}{c}{\textbf{Pair 3}} & \multicolumn{2}{c}{\textbf{Pair 4}} \\
\cmidrule(lr){2-3}\cmidrule(lr){4-5}\cmidrule(lr){6-7}\cmidrule(lr){8-9}
\textbf{Train} & \multicolumn{2}{c}{FLIVE} & \multicolumn{2}{c}{FLIVE} & \multicolumn{2}{c}{KonIQ10K} & \multicolumn{2}{c}{CLIVE} \\
\textbf{Test}  & \multicolumn{2}{c}{CLIVE} & \multicolumn{2}{c}{KonIQ10K} & \multicolumn{2}{c}{CLIVE}   & \multicolumn{2}{c}{KonIQ10K} \\
\cmidrule(lr){2-3}\cmidrule(lr){4-5}\cmidrule(lr){6-7}\cmidrule(lr){8-9}
\textbf{Metric} & SRCC & PLCC & SRCC & PLCC & SRCC & PLCC & SRCC & PLCC \\
\midrule
SigLIP2                                   & 0.0133 & 0.0226 & 0.0292 & 0.0236 & 0.0071     & 0.01182     & 0.0015     & 0.0021     \\
SigLIP2 + Sigmoid                         & 0.0096 & 0.0058 & 0.0015 & 0.0008 & 0.0278     & 0.0236     & 0.0019     & 0.0011     \\
SigLIP2 + Activation Gating        & 0.0249 & 0.0168 & 0.0034 & 0.0071 & 0.0013    & 0.0018     & 0.0050     & 0.0009     \\
\bottomrule
\end{tabular}
\end{table*}


\begin{table*}[!htbp]
\centering
\caption{Compute profile for SigLIP2 variants. All counts are single-image (batch=1) unless noted.}
\label{tab:compute_siglip2}
\footnotesize
\setlength{\tabcolsep}{4pt}
\begin{tabular}{@{}lccccccccccc@{}}
\toprule
Variant & Input (H$\times$W) & Patch & MACs (G) & Mem (MiB) & Batch & Precision & Framework & Counter & Attn inc. \\
\midrule
SigLIP2 +(LReLU) & 512$\times$512 & 16 & \emph{423.99}  & \emph{2270} & 1 & bf16 & PyTorch~2.1 & fvcore & Yes \\
SigLIP2 + Sigmoid    & 512$\times$512 & 16 & \emph{423.99}  & \emph{2270} & 1 & bf16 & PyTorch~2.1 & fvcore & Yes \\
SigLIP2 + Activation Gating       & 512$\times$512 & 16 & \emph{423.66}  & \emph{2270} & 1 & bf16 & PyTorch~2.1 & fvcore & Yes \\
\bottomrule
\end{tabular}

\vspace{3pt}
\begin{minipage}{0.96\linewidth}\footnotesize
\end{minipage}
\end{table*}


\begin{figure*}[t]  
  \centering
  \includegraphics[width=.78\linewidth]{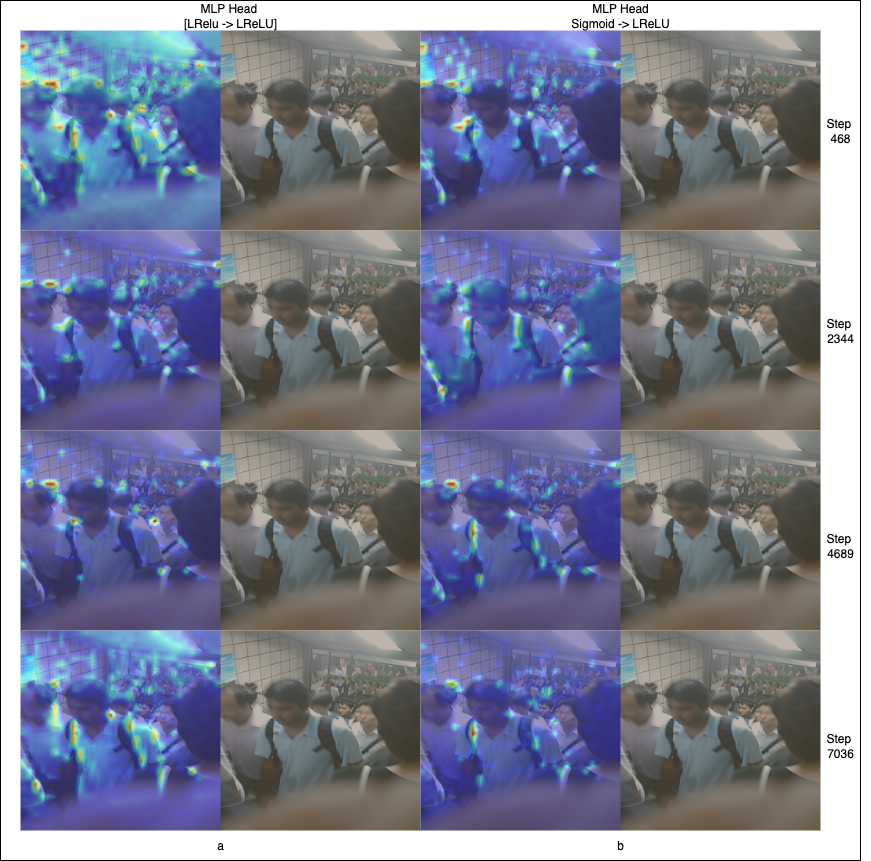}
  \caption{Grad-CAM visualizations comparing an MLP head without sigmoid (a) versus with sigmoid gating (b) over training steps. Introducing the sigmoid reduces the dominance of high-activation (strongly semantic) channels in the backbone features early in training, promoting reliance on medium-response evidence. This yields a clearer correlation with the facial blur artifact in (b), while the baseline in (a) frequently attends to unrelated salient structure and fails to highlight the facial blur.}
  \label{fig: Grad-CAM SIGLIP MSE vs SIGLIP SIGMOID}
\end{figure*}



\begin{figure*}[t]  
  \centering
  \includegraphics[width=.75\linewidth]{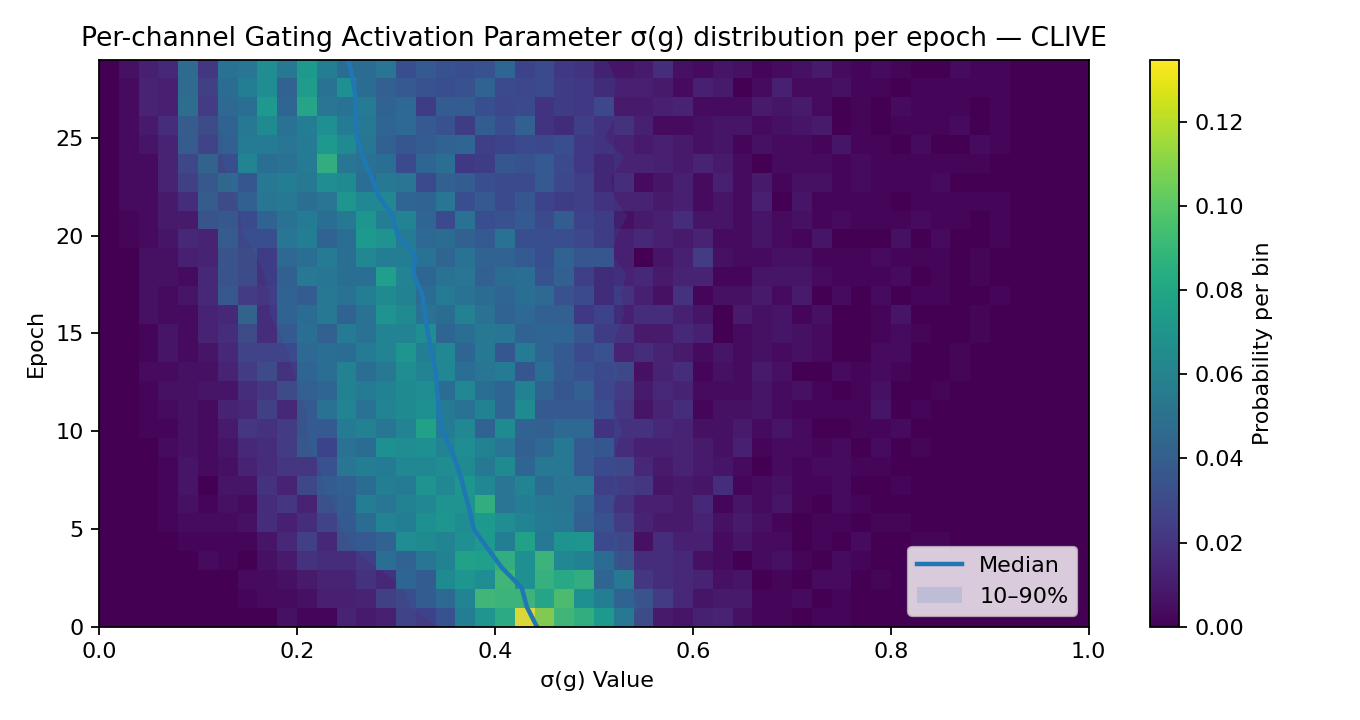}\hfill
  \includegraphics[width=.75\linewidth]{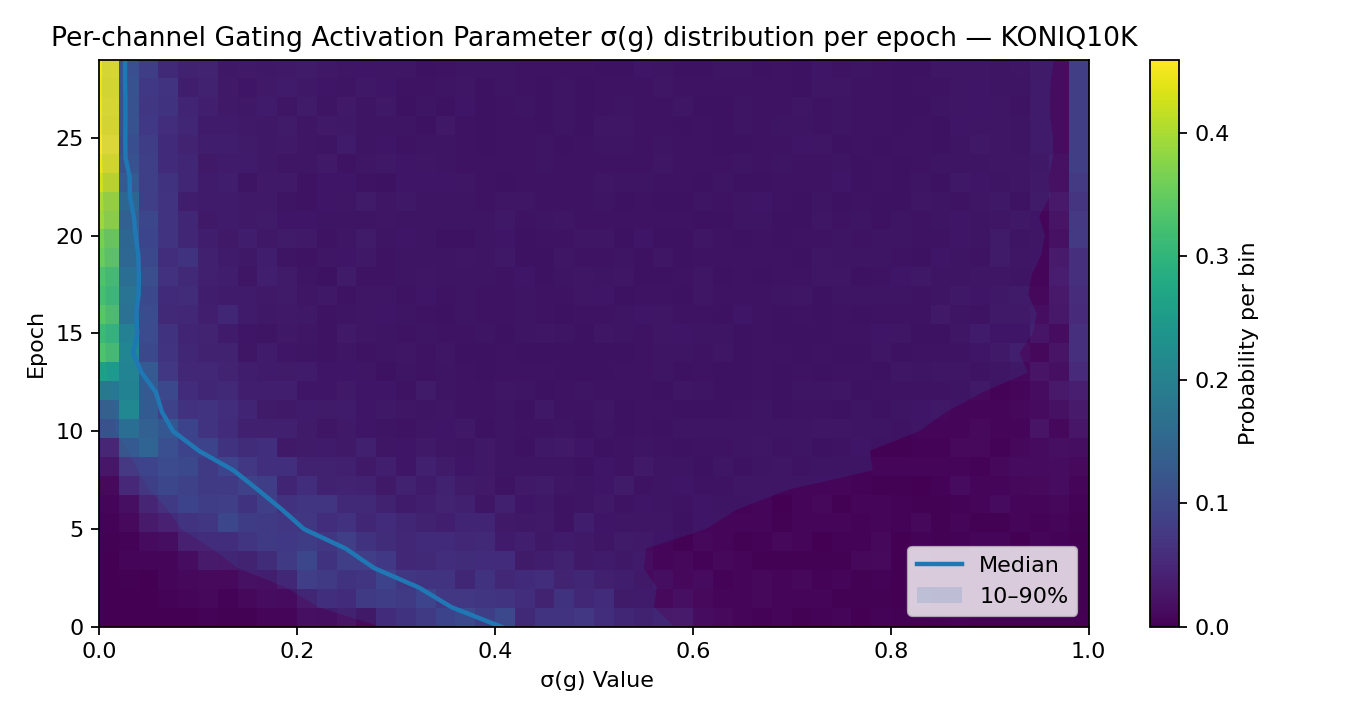}
  \caption{Channel-wise distributions of the gate weight $w=\sigma(g)$learned by the gated activation head across different epochs, comparing CLIVE (low-data regime) and KonIQ-10k (large-data regime). Larger w indicates greater reliance on the sigmoid branch, while smaller w favors the LeakyReLU branch.}
  \label{fig: channel_wise_distirbution_gating_MLP_per_epoch}
\end{figure*}


\end{document}